# Complexity measurement of natural and artificial languages

Gerardo Febres[1], Klaus Jaffé[1], Carlos Gershenson[2,3]

[1] *Laboratorio de Evolución, Universidad Simón Bolívar, Venezuela.*
[2] *Instituto de Investigaciones en Matemáticas Aplicadas y en Sistemas, Universidad Nacional Autónoma de México*
[3] *Centro de Ciencias de la Complejidad, Universidad Nacional Autónoma de México*

**Abstract**
*We compared entropy for texts written in natural languages (English, Spanish) and artificial languages (computer software) based on a simple expression for the entropy as a function of message length and specific word diversity. Code text written in artificial languages showed higher entropy than text of similar length expressed in natural languages. Spanish texts exhibit more symbolic diversity than English ones. Results showed that algorithms based on complexity measures differentiate artificial from natural languages, and that text analysis based on complexity measures allows the unveiling of important aspects of their nature. We propose specific expressions to examine entropy related aspects of tests and estimate the values of entropy, emergence, self-organization and complexity based on specific diversity and message length.*

## 1 Introduction

The study of symbol frequency distribution for English was initially addressed by Zipf [1] in 1949 and Heaps during the 70s [2] , giving rise to Zipf's and Herdan-Heaps' laws respectively (frequently referred to as Heaps' law). Zipf [1] suggested that the scale free shape of the word frequency distribution, typically found for English long texts, derives from his *Principle of Least Effort*. As in many other large scale phenomena, the origin of the tendency of natural languages to organize around scale free structures, remains controversial [3] and a plentiful source of hypothesis and comparisons with other 'laws of nature' [4] [5] [6]. The relationship between both Laws has been studied [7] and their validity for various natural alphabetic languages tested [8] [9] [10]. Yet, a generally accepted mechanism to explain this behavior is still lacking, as Zipf's and Heaps' laws have been traditionally applied only to probabilistic consequences of grammar structure and language size.

Language grammar has been addressed in the study of basic grammar rules and the mechanisms to buildup English phrases, initiated by Chomsky [11] in the late 50's. Later Jackendoff developed the X-bar theory [12], fostering the idea of underlying effects driving human communication processes to produce grammar properties common to all natural languages. Yet clear descriptions of the fundamental sources of such a behavior, remains a matter of discussion, perhaps because it is a problem too complex to be completely understood employing only theoretical methods.

In this paper, we compare messages expressed in natural and artificial languages using metrics developed to quantify complexity. Our comparison is based on measurements of message symbol diversity, entropy and symbol frequency distributions. Zipf's distribution profiles and Heaps' functions are identified for different messages samples. We evaluate the impact of these measures over emergence, self-organization and complexity of messages expressed in natural and artificial languages.

Our strategy is to evaluate a wide range of texts for each language studied, including text pieces from a variety of writers distributed over a timespan of more than 200 years. All texts were recorded in a computer file directory and analyzed with purposely developed software called *MoNet* [13] (see section 2.10), as explained in Sections 2.1 to 2.6.

# 2 Methods

We compared three aspects of English, Spanish and artificial languages: symbol diversity $D$, entropy $h$, and the symbol frequency distribution $f$. For the available measures of diversity and information, we follow Gershenson and Fernandez [14] to evaluate emergence and self-organization for natural and artificial languages. For complexity, we use the definition of Lopez-Ruiz et al. [15], which sees complexity as a balance between chaotic and stable regimes.

## 2.1 Text length $L$ and symbolic diversity $d$

The length of a text $L$ is measured as the total number of symbols or words used and the diversity $D$ as the number of different symbols that appear in the text. We define the specific diversity $d$ as the ratio of diversity $D$ and length $L$ , that is

$$d = \text{specific diversity} = D/L \,. \tag{1}$$

In this study symbols are considered at the scale of words. Here a word is a considered as a sequence of characters delimited by some specific characters such as a blank space (see section 2.9). Most recognized symbols were natural and artificial language words. Nevertheless some single character symbols, such as periods and commas, appeared by themselves with complete meaning and function and therefore playing a role comparable to that of normal words.

## 2.2 Entropy $h$

Entropy calculations are based on Shannon's information [16], which is equivalent to Boltzmann-Gibbs entropy. Message information is estimated by the entropy equation is based on the probability of appearance of symbols within the message. Symbols (words) are treated all with the same weight, ignoring any information that might be associated to meanings, length or context. Shannon's entropy expression for a text with a symbol probability distribution $P(p_i)$ is:

$$h(P(p_i)) = -\sum_{i=1}^{2} p_i \; log_2 \; p_i \,. \tag{2}$$

Shannon was interested in evaluating the amount of information and its transmission processes; therefore his entropy expression was presented for a binary alphabet formed by the symbols '0' and '1'. Entropy measurement in this study is at the scale of words, where each word is a symbol, extending the original Shannon's expression for a D-symbol alphabet:

$$h(F(f_r)) = -\sum_{r=1}^{D} \frac{f_r}{L} \log_D \frac{f_r}{L} \,, \tag{3}$$

where we have replaced the term $P(p_i)$ with its equivalent in terms of the symbol frequency distribution $F(f_r)$ and the text length $L$ measured as the total number of symbols. The values for the symbol frequency distribution $F(f_r)$ are ordered on $r$, the symbol rank place ordered by their number of appearances in the text. Since there are $D$ different symbols, $r$ takes integer values from 1 to $D$. Notice the base of the logarithm is the diversity $D$ and hence $h$ is bounded between zero and one.

## 2.3 Emergence *e*

As a system description is based on different scales—the number of different symbols used—the quantity of information of the description varies. Emergence measures the variation of the quantity of information needed to describe a system as the scale of the description varies, thus, emergence can be seen as a profile of quantity of information for a range of system scales. Therefore we express emergence $e$ as a function of the quantity of information respect to the description length $L$ (total number of symbols) and the specific symbol diversity $d$. This is given Shannon's information (3), so we have:

$$\mathrm{e}(F(f_r)) = h(F(f_r)) \,. \tag{4}$$

## 2.4 Self-organization *s*

The self-organization of a system can be seen as the capacity to spontaneously limit the tendency of its components to fill the system space—symbols in our case—in a homogenous, totally random distributed fashion. Since entropy reaches a maximum when the system components are homogeneously randomly dispersed, self-organization $s$ is measured as the difference of the maximum entropy level $h_{max} = 1$, and the actual system entropy [17].

$$s(F(f_r)) = h_{max} - h(F(f_r)) = 1 - e(F(f_r)) \,. \tag{5}$$

## 2.5 Complexity *c*

Message entropy calculations are based on Shannon's expression [16]. Message information is estimated by the entropy equation based on the probability of appearance of symbols within the message. Symbols (words) have all the same weight here, ignoring putative differences in information associated to the word's meanings, length or context. We used the complexity definition proposed by López-Ruiz et al. [15], and its quantifying expression proposed by Fernández et al. [17]

$$c(F(f_r)) = 4 \cdot e(F(f_r)) \cdot s(F(f_r)) = 4 \cdot h(F(f_r)) \cdot [1 - h(F(f_r))] \,. \tag{6}$$

In this definition, complexity is high when there is a balance between emergence (entropy, chaos) and self-organization (order). If either is maximal, then complexity is minimal.

## 2.6 Symbol frequency distribution *f*

For any message or text the number of words in a rank segment $[a, b]$ may computed as:

$$L_{a,b} = \sum_{r=a}^{b} f_r \,, \tag{7}$$

where $a$ and $b$ are respectively the start and the end of the segment where symbol were ranked. For any segment, a=1 and b=D.

Zipf's law states that any sufficiently long English text will behave according to the following rule [3] [8]:

$$f(r) = \frac{f_a}{(r-a)^g}, \tag{8}$$

where $r$ is the ranking by number of appearances of a symbol, $f(r)$ a function that retrieves the numbers of appearances of word ranked as $r$, $f_a$ the number of appearances of the first ranked word within the segment considered, and $g$ a positive real exponent.

For any message, we define Zipf's reference $Z_{a,b}$ as the total number of symbol appearances in the ranking segment [a, b] assuming that it follows Zipf's Law. Therefore $Z_{a,b}$ is

$$Z_{a,b} = \sum_{r=a}^{b} f_r = \sum_{r=a}^{b} \frac{f_a}{r^g}. \tag{9}$$

Eq. (8) allows us to determine the Zipf's reference $Z$ for any segment within the symbol rank dominion. We computed versions of Zipf's reference $Z$ for the complete message, specifically named $Z_{1,D}$, and for the tail of the message frequency distribution (see Figure 1), named $Z_{\theta,D}$. The sub index $\theta$ is used to indicate the ranking position $r_\theta$ where the head-tail transition occurs.

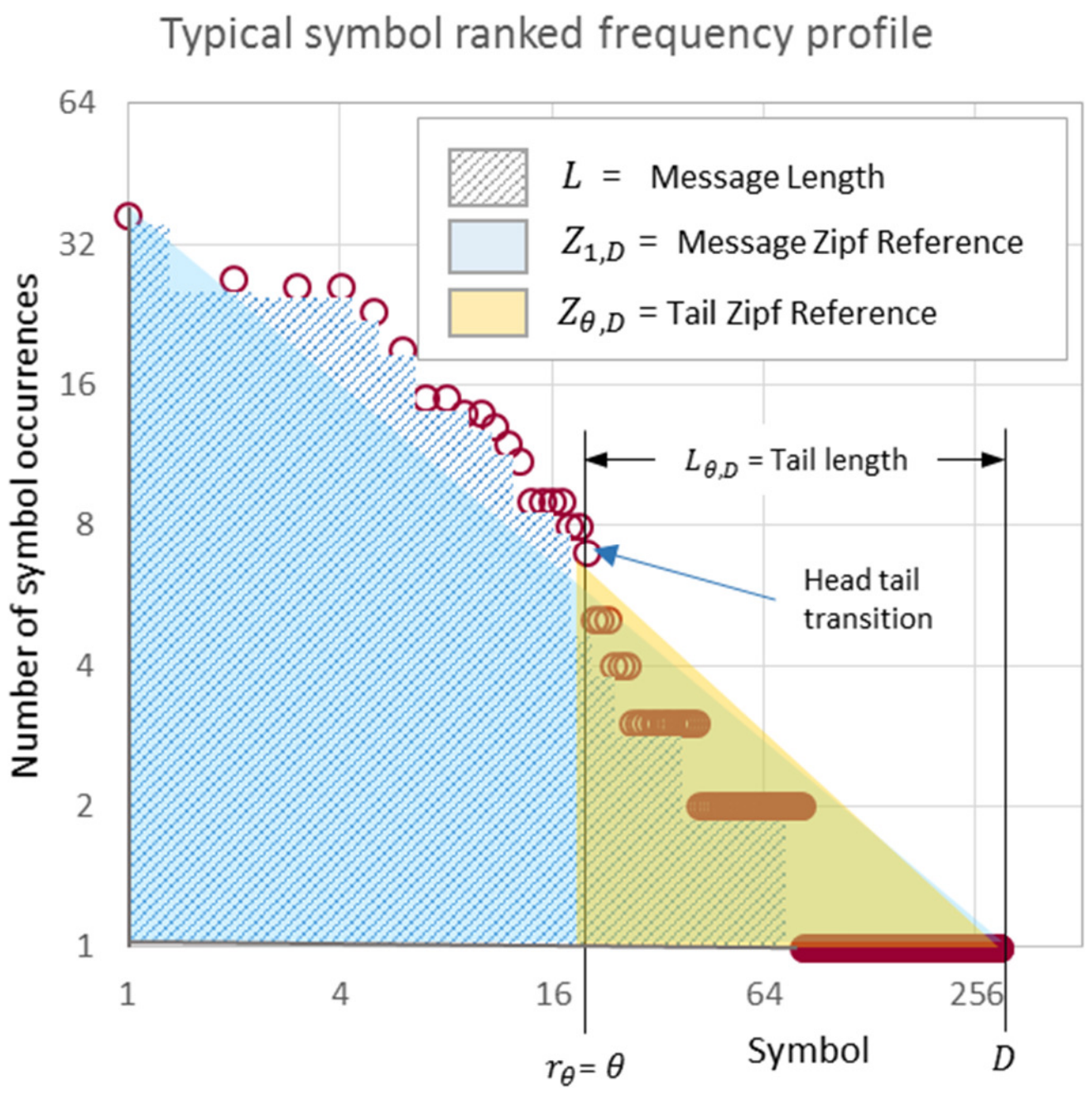

Figure 1. Typical symbol ranked profile. Red dots indicate the number of occurrences and the ranking position of the symbols of a given text. Message Zipf's and tail Zipf's references are the blue and yellow shadowed areas respectively.

Head-tail transition location can be a difficult parameter to set and is often considered to be among a range of possibilities. We used the following definition: For a discrete symbol ranked frequency or probability distribution, the region of the lowest frequency of ranked symbols starts where the symbols with a unique frequency (or probability = 1) end. Figure 1 illustrates an example of symbol frequency profile. The point signaled with the arrow corresponds to the 20th rank position and has 7 occurrences, and no other symbol shares the same number of appearances. At that point we define the start of the tail which includes the distribution domain shadowed in yellow in the figure.

### 2.7 Zipf's deviation $J$ for a ranked distribution

The complete message Zipf's reference is determined by expression (8). The corresponding Zipf's deviations $J_{1,D}$ from a Zipfian distribution and the deviation of its tail $J_{\theta,D}$ are

$$J_{1,D} = {(L - Z_{1,D})}/{Z_{1,D}} , \quad J_{\theta,D} = {(L_{\theta,D} - Z_{\theta,D})}/{Z_{\theta,D}} . \qquad \text{(10a)(10b)}$$

Identifying the starting point for the tail of each message or code profile is a search intensive task. We included in the software *MoNet*, the capability of locating within a frequency profile the points with properties characterizing the start of the tail, and to split messages and codes in heads and tails. Once the tail starting rank $r_\theta$ is determined, Zipf's tail deviation was obtained by applying equations (10a) and (10b).

## 2.8 Message selection

We built text libraries containing consisting of large text fragments, obtained from English and Spanish speeches, segments of stories and novels, and computer codes written in high level programming languages (C, C#, Basic, Matlab, Java, HTML and PHP). The program then produced descriptive indices and attributes for each of these. Each message could be analyzed as an individual object or as a part of a collective group of objects.

**Natural language message selection**

Natural language messages were collected from historic speeches available in on the web as texts expressed in English or Spanish. Natural language texts include speeches from politicians, human rights defenders and literature Nobel Laureates. The language used to write the original speech was not a selection criterion. There are speeches in our selection originally written in English, Spanish, French, Russian, Italian, German, Arabic, Portuguese, Chinese and Japanese. Translated speeches and texts are indicated as such, providing data for studying translations. Novel fragments were authored in English or Spanish by popular writers and by some Nobel laureates in literature.

We collected 156 texts in English and 158 in Spanish. The shortest speech was 87 words long, whereas the longest speech contained more than 20000 words.

**Artificial language message selection**

We included 49 computer codes devoted to perform recognizable tasks. Artificial text lengths go from a C# code which generates Fibonacci numbers with just 62 symbols, to computer logs with more than 160000 symbols. This selection of artificial texts include codes written in *C*, *C#*, Basic, Java, *MatLab*, *HTML* and *PHP*. The Table in Appendix A gives details of codes and their fragments used here.

## 2.9 Symbol treatment

Special treatment of certain character strings or symbols were considered as follows:

**Word**: A word is any character string isolated by the characters 'space' or 'line return'. The word is the symbolic unit.
**Space**: The space works as a delimiter for symbols or words.
**Line Return or Line Feed**: Is a delimiter for paragraphs.

**Punctuation Signs**: Any sign is considered as a complete independent symbol. In natural languages, the punctuation signs have specific meaning that, with very few exceptions, are not sensitive to other surrounding characters. When located next to numeric characters, if a punctuation sign appeared attached to another symbol, the sign was handled as being separated by the space character to keep it as a single symbol. This rule provides a coherent solution to the very frequent case where words appear attached to punctuation symbols.
**Numbers**: For natural languages, a digitally written number might be a unique sequence of characters. Numbers express quantities and work as adjectives or modifiers of an idea. All numbers in a natural language message are then considered as different symbols.
**Synonyms**: Since ours is a symbolic analysis, synonyms are considered as different words.
**Capital letters**: Words are case sensitive. In English and Spanish a word with its first letter written with a capital letter, refers to a specific name. Therefore, a name appropriately written with a first capital letter is different from the same character sequence written with all letters in lower case. But when the word starting with capital letter comes after a period sign, we assume it is a common lower case word, unless other appearances of the same word indicates it certainly is a proper name that should keep its first capital letter.

**For Spanish messages**:
**Accents**: in Spanish, vowels are sometimes marked with an accent over it to indicate where the sound stress or emphasis should be. Rules to indicate when the accent mark should be present and when it shouldn't, are easy to apply and are part of what any Spanish speaker should know from elementary school. Forgetting accent marks when they should appear is associated with poor writing abilities; it is unacceptable in any serious literary work. We consider that any accented word is different, and has some different meaning, from the same character sequence without accents.

**For artificial languages (computer code)**:
**Comments**: in artificial languages comments do not affect any action of the interpreter or compiler. Additionally, comments are intended to convey ideas to the human programmer, administrator or maintenance personnel, hence most comments are written in phrases dominated by natural languages. Comments were thus excluded from any code analyzed.
**Computer Messages**: Most computer codes rely on the possibility of informing the user or operator about execution parameters. This information is normally expressed in different languages to that of the code. Computer message contained in a code were converted to a single word by extracting all spaces.
**Numbers**: Differently from natural languages, in artificial languages sequences of digits may represent variable names or memory addresses, which are objects with different meaning. In artificial languages, any difference in a digit is considered to result in a different word.
**Capital letters**: We considered artificial language symbols as case sensitive.
**Variables**: When in different parts of the code, two or more variable names were presented as the same symbol or characters string, but we know that sometimes they could have a totally different meaning since they could be pointing to a different memory address. This may introduce some deviation in the results.

## 2.10 Software
Two software programs were developed to analyze the texts. First, we built a file directory structure containing, and classifying the messages each with its inherent and invariant text-object properties. We refer to the file directory as the library. The second software program, called *MoNet*, manage the library and produced the data for our study.
**Library**

The library holds descriptions of each existing text-object with its attribute values. The scope of each object description can be adjusted adding attributes or even modifying their data representation nature and dimensionality. We built a text library containing hundreds of these text-objects. Libraries can be updated by deleting or adding text-objects.

***MoNet***

*MoNet* is a bundle of scripts, interpretations, programs and visual interfaces designed to analyze complex systems descriptions at different scales of observation. *MoNet* describes a system as a collection of objects and object families connected by hierarchical and functional relationships.

The scope of each object description can be adjusted adding attributes or modifying their representation and dimensionality. *MoNet* can treat every text included in a library as well as the library itself, offering results for text-objects as independent elements or as groups. For every component of the system modeled, descriptions at different scales can co-exist. Individual objects can be selected combining logical conditions based on properties or attribute values.

# 3 Results

## 3.1 Diversity for natural and artificial languages

Figure 2 shows how diversity varies with the message length in texts written in English, Spanish and Computer Code. Diversity increases as messages grow in length, but there seems to be an upper bound of diversity for each message length. For English this upper bound is slightly lower than for Spanish. As message length increases, English also shows a wider dispersion toward lower diversities of words. Artificial messages represented by computer code showed a much lower diversity than the natural languages. The regression models of Heaps' Law [9] for message diversities and message length are:

$$English: \quad D = 3.766 \cdot L^{0.67} \,. \tag{11a}$$

$$Spanish: \quad D = 2.3 \cdot L^{0.75} \,. \tag{11b}$$

$$Software: \quad D = 2.252 \cdot L^{0.61} \,. \tag{11c}$$

## 3.2 Entropy for natural and artificial languages

Figure 3 shows entropy $h$ values for texts expressed in natural languages and computer code programs as a function of specific diversity $d$ (see section 2.1). Extreme values of entropy are the same for messages expressed in all languages; entropy drops down to zero when diversity decreases to zero and tends to a maximum value of 1 as specific diversity approaches 1. For artificial messages entropy is dispersed over a wider range of values, perhaps as a consequence of the many different computer languages included in this work's sample. Natural languages show less dispersion in entropy levels, nevertheless differences among languages show up in the areas they cover over the plane of entropy-diversity with few overlapping shared areas over that space. See Figure 3.

The entropy expression shown in Eq. (3) is a function with $D - 1$ degrees of freedom; there are $D - 1$ different ways of varying the variable $F$ that affect the resulting value of entropy $h$. Nevertheless, when specific diversity is at extreme values $d = 0$ and $d = 1$, the distribution $F$ becomes homogenous and function $h(F)$ adopts the following predictable behavior.

1. $h(F \mid d \to 0) = 0 \,.$ (12a)
2. $h(F \mid d \to 1) = 1 \,.$ (12b)

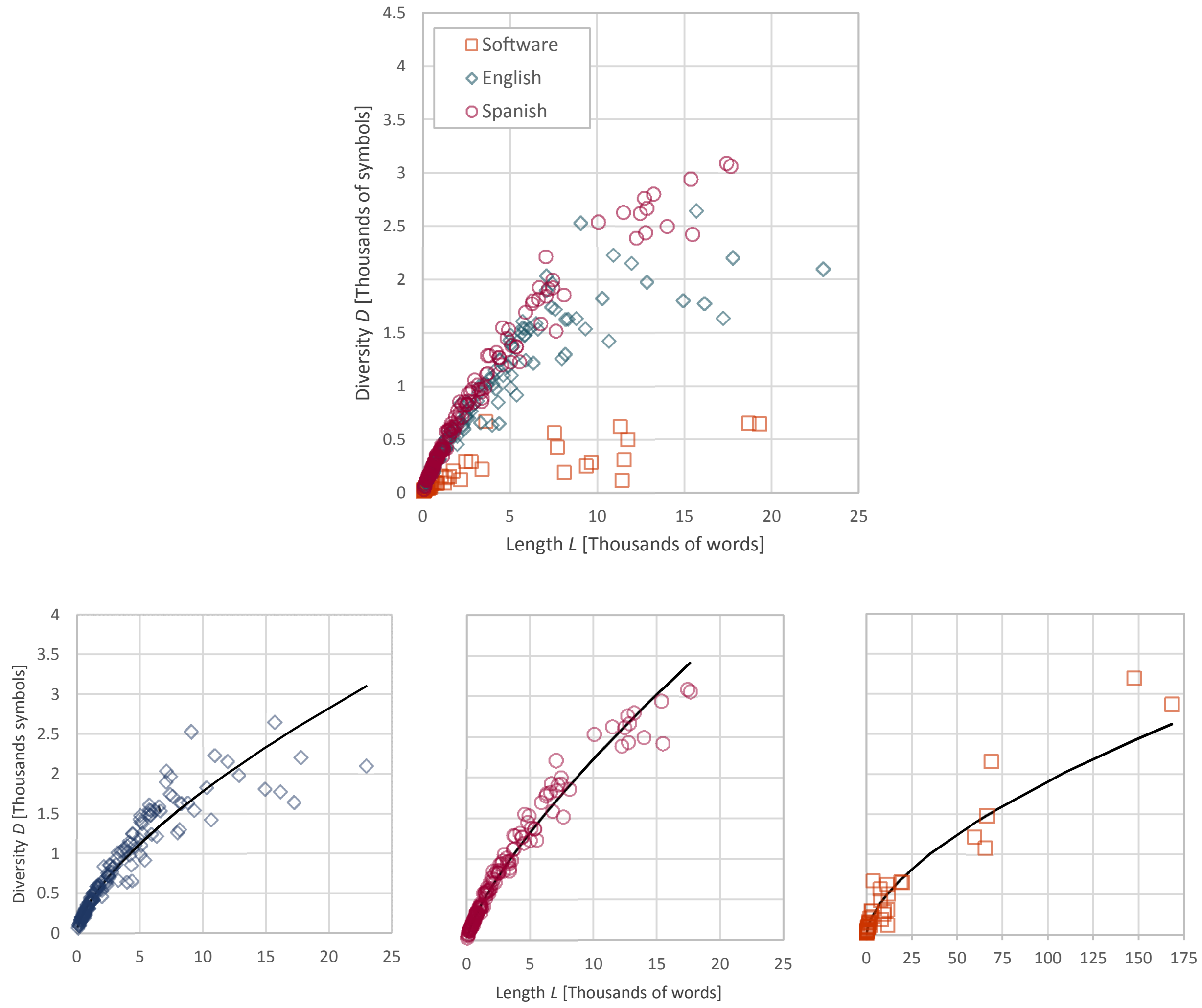

Figure 2. Diversity for messages expressed in English, Spanish and Computer Code. Lower row presents fit dots (black) for messages expressed in English (left), Spanish (center) and Software (right).

Having these extreme conditions for $h(F)$ , we propose a real function $h(d)$ to characterize the entropy distribution of a language over the range of specific diversity. The dispersion of the points is due to the fact that none of the texts obeys perfectly a Zipf's law, yet each language tends to fill a particular area of the space entropy-specific diversity. To model the curves along the core of these clusters of dots, that is entropy as a function of specific diversity, we refer to the so called Lorenz curves [18] which can be used to describe the fraction of edges $W$ of a scale-free network with one or two ends connected to a node which belongs to the fraction $P$ of the nodes with highest degree [5]. The family of Lorentz curves is expressed by

$$W = P^{(\alpha-2)/(\alpha-1)} . \qquad (13)$$

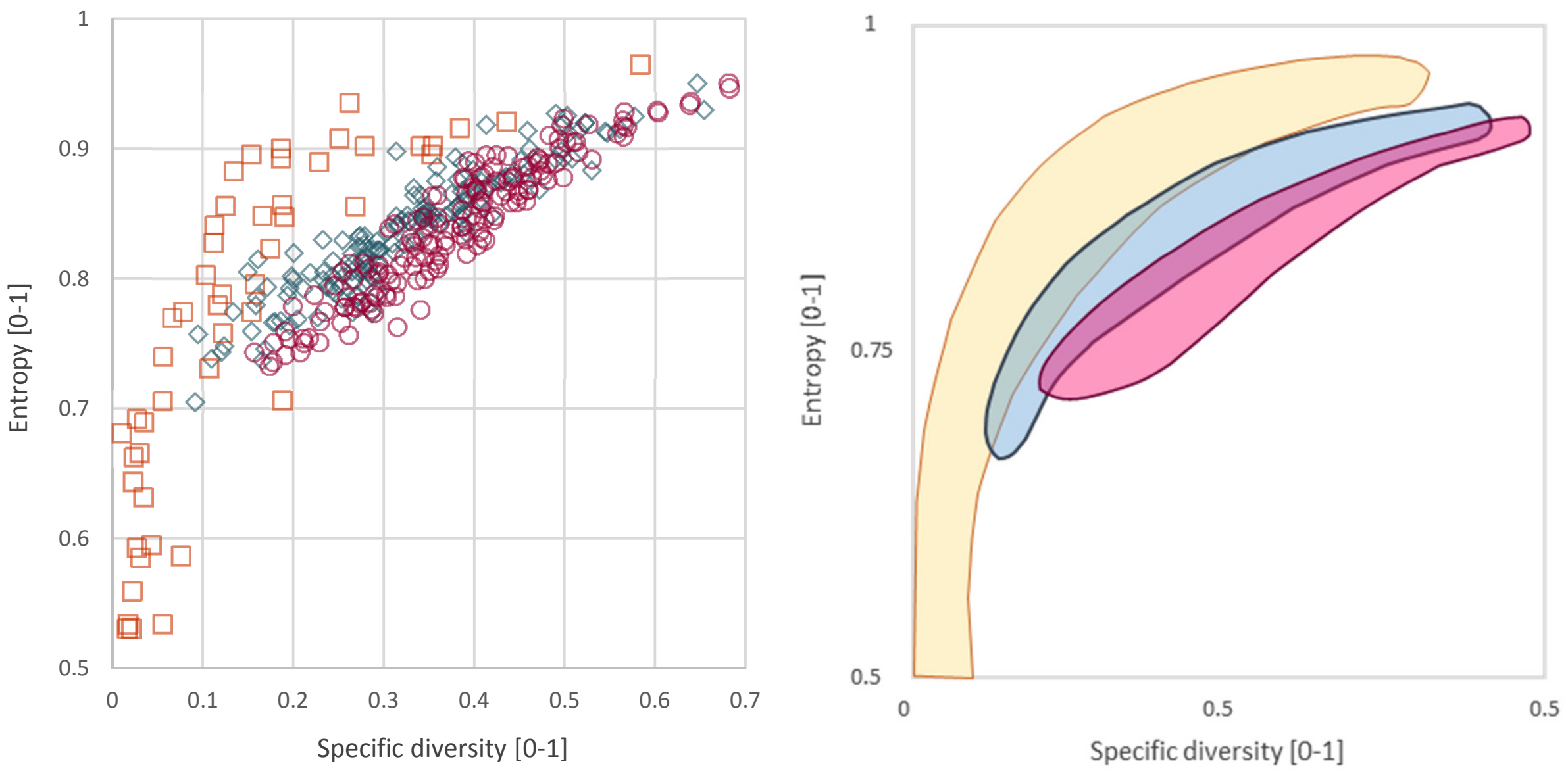

Figure 3. Messages entropy vs specific diversity for English, Spanish and computer code. On the left graph each dot represents a message. The right graph shows the area where most messages lie upon its corresponding language.

Now consider the network associated to a text where the nodes represent words or symbols and the edges represent the relation between consecutive words. In a network like this, all nodes, except those corresponding to the first and the last words, will have a degree of connectivity that doubles the number of appearances of the represented word. Thus, the resulting ranked node degree distribution will be analogous to a Zipf's distribution and therefore, the network as defined, will have a scale-free structure. On the other hand, entropy can be interpreted as the cumulative uncertainties that every symbol adds or subtracts from the total uncertainty or entropy. Viewing entropy $h$ of a ranked frequency distribution as the cumulative uncertainty after adding up the contributions of the $D$ most frequent symbols, we should expect this entropy $h$ to have a scale-free behavior with respect to changes $D$. After the analogies between these conditions and those needed to expect a behavior like the Lorentz curves dictate, we propose the use of the one-parameter expression (13) to describe any language's entropy as a function of $d$ and the parameter$\alpha$. So that:

$$h = \left(\frac{D}{L}\right)^{(\alpha-2)/(\alpha-1)} = d^{(\alpha-2)/(\alpha-1)} \;. \qquad (14)$$

Figure 4 compares the data using the entropy model for the languages studied. Values of $\alpha$ were obtained to minimize square errors between the entropy model and the experimental results obtained from each text of the library. Numerical results were $\alpha = 2.123$ for English, $\alpha = 2.178$ for Spanish and $\alpha = 2.1$ for artificial. The figure shows a much wider range of entropy values for artificial languages compared to the natural languages studied. Equations (15a), (15b), and (15c) present specific cases of function $h(d)$ for each language studied:

$$English: \quad h = d^{0.1511} \qquad (15a)$$

$$Spanish: \quad h = d^{0.1756} \qquad (15b)$$

$$Software: \quad h = d^{0.09091} \qquad (15c)$$

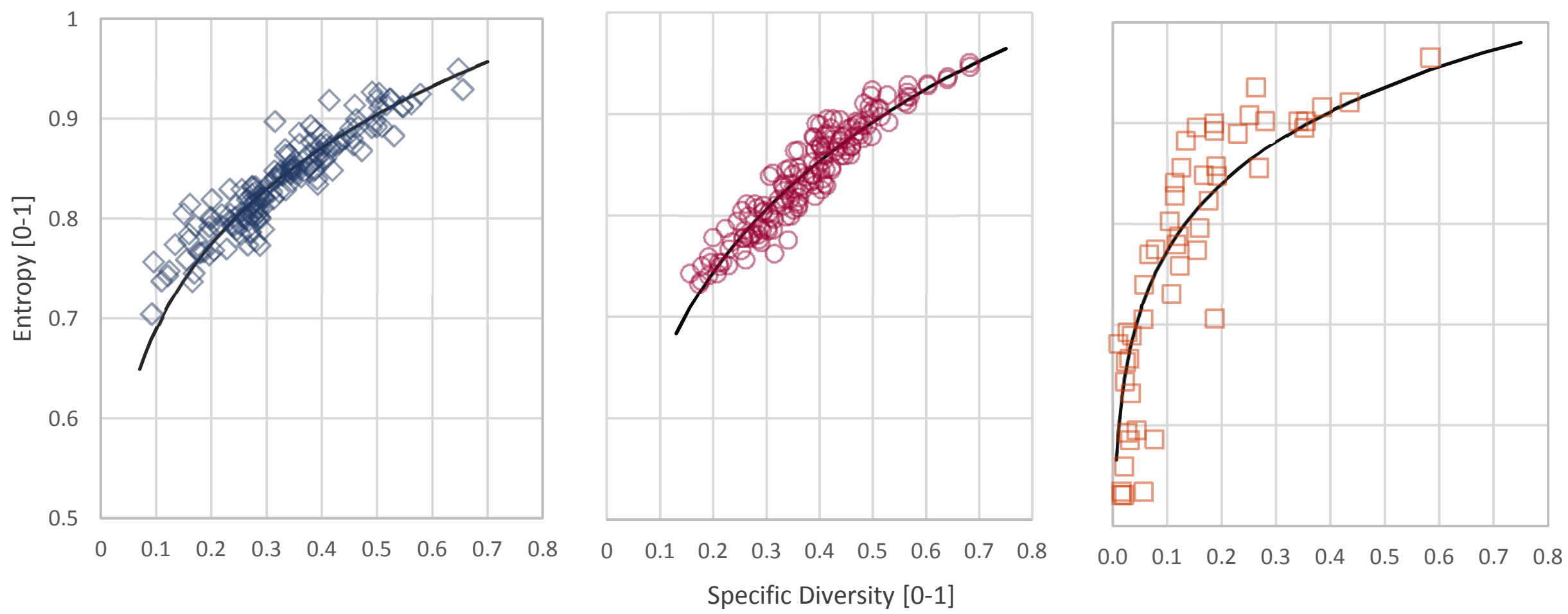

Figure 4. Messages entropy vs specific diversity for English (left), Spanish (Center) and artificial languages (right).

## 3.3 Emergence, self-organization and complexity

Starting with functions for entropy, obtaining expressions for emergence, self-organization and complexity is straightforward using results of Equations (15a), (15b) and (15c) with Equations (4), (5) and (6). Figure 5 illustrates these results.

To obtain expressions of emergence, self-organization as functions of the message length $L$, we combined equations (15a), (15b) and (15c) with (11a), (11b) and (11c) respectively. See the results in Figure 6.

For all languages, emergence increases with specific diversity and decreases with length. Self-organization follows opposite tendencies, decreasing with specific diversity and increasing with length. Complexity is maximal for low specific diversities and then decreases, although much less for natural languages. Complexity increases with length for all languages.

The most conspicuous result here is that artificial languages show a different pattern in complexity depending on specific diversity, as the maximum complexity for artificial languages is close to zero and then decreases faster than natural languages. This might reflect fundamental differences in organizing the symbols (grammar) between both types of languages.

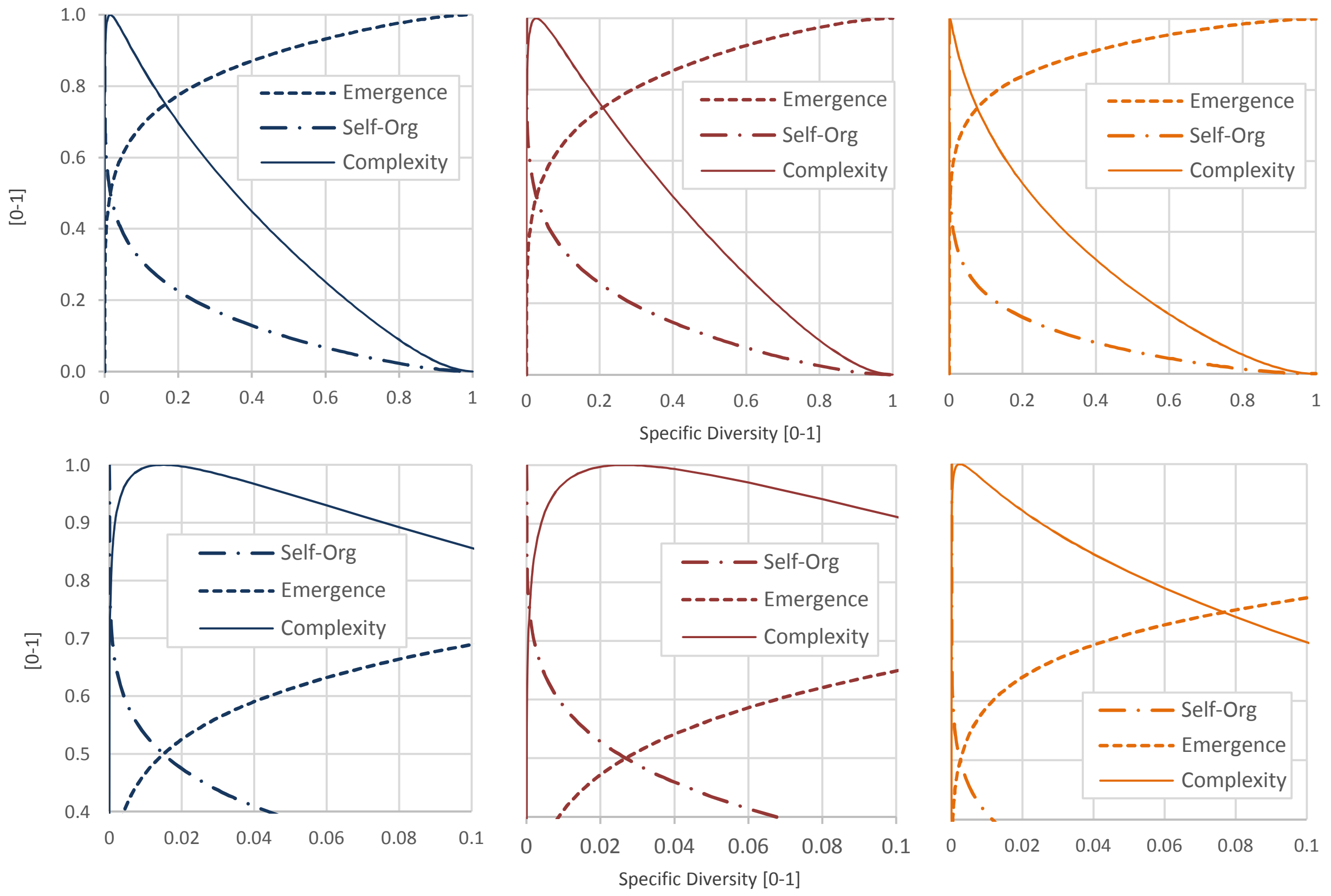

Figure 5. Emergence, self-organization and complexity for English (left), Spanish (Center) and Computer Code (right). Vertical axis is dimensionless [0-1]. Graphs placed on the lower row correspond to the detail very near the value zero for horizontal axis. These plots are based on Equations (4), (5) and (6) combined with Equations (15a), (15b) and 15c).

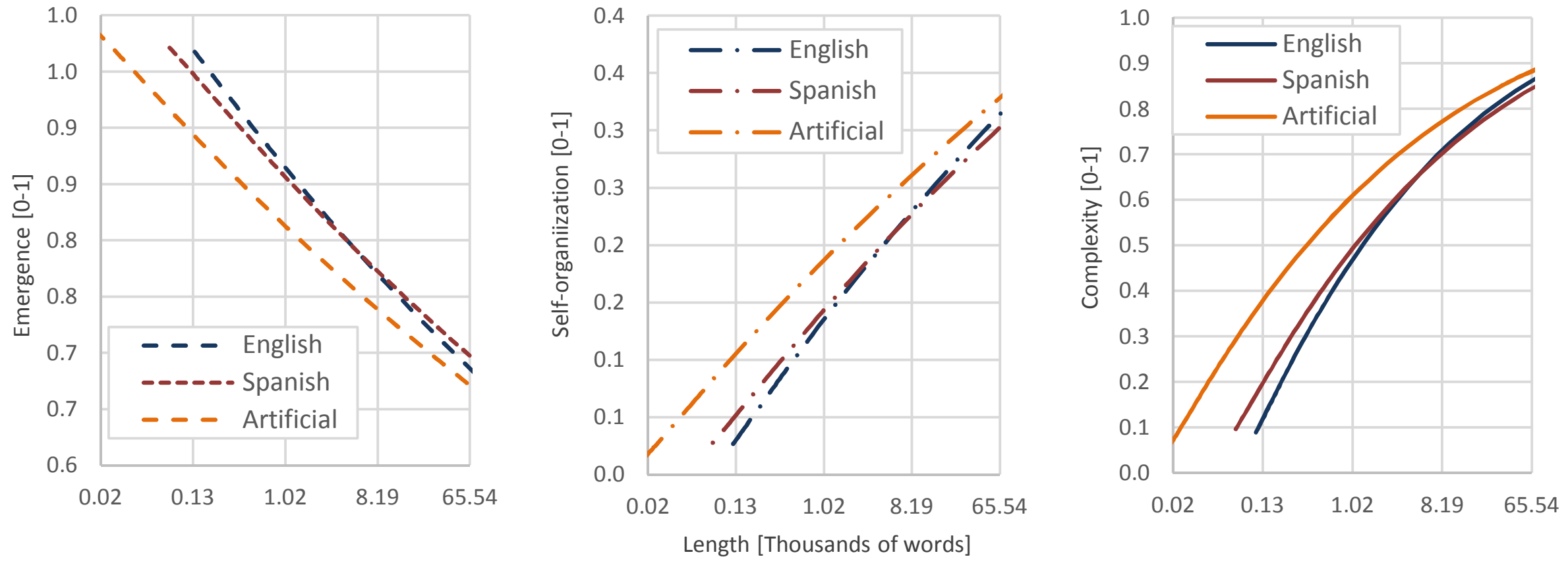

Figure 6. Emergence (left), self-organization (center) and complexity (right) for English, Spanish and Computer Code. Vertical axis are dimensionless [0-1]. . These plots are based on Equations (4), (5) and (6) combined with Equations (15a), (15b), (15c), (11a), (11b) and (11c).

## 3.4 Symbol Frequency distributions

Profile of symbol frequency distributions were inspected in two ways: first by a qualitative analysis of their shapes and second by characterizing each profile with its area deviation $J$ with respect to a Zipfs distributed profile.

A sample of symbol frequency distributions profiles for the considered languages is represented in Figure 7. Each sequence of markers belongs to a message and each marker corresponds to a word or symbol within the message. While no important differences are observed among messages profiles expressed in the same language, a noticeable tendency to a faster decreasing frequency profile appears for messages expressed in artificial languages, perhaps a consequence of the limited number of symbols these types of languages have.

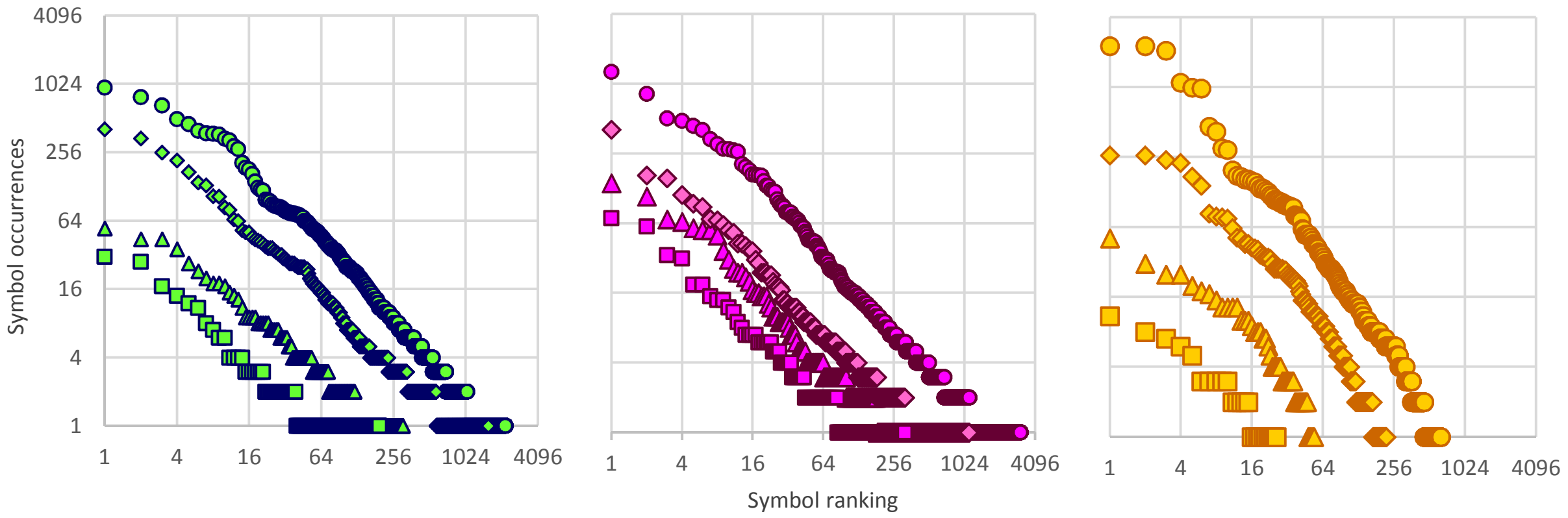

Figure 7. Ranked symbol frequency distribution for English (left), Spanish (center) and Computer Code (right). A sample of three or four messages for each language is shown. English: square -*1945.BS.Eng.GabrielaMistral*, triangle - *1921.MarieCurie,* rhombus - *1950.NL.Eng.BertrandRussell,* circle - *1890.RusselConwell.* Spanish*:* square - *1936.DoloresIbarruri*, triangle *1982.Gabriel García Márquez* - rhombus - *JoseSaramago.Valencia,* circle - *CamiloJoseCela.LaColmena.Cap1.* Artificial*:* square - *FibonacciNumbers.CSharp,* triangle - *QuickSort.CSharp,* rhombus – *Sociodynamica.Module3,* circle - *WebSite.Inmogal.php.*

By building these frequency profiles, we could obtain a list of the most used words in English and Spanish. An equivalent list for artificial languages is also obtainable; however it is difficult to interpret due to the diversity of programming languages used in our artificial text sample. Table 1 shows statistics about the use of symbols for English and Spanish. Table 1 was constructed overlapping symbol frequency profiles of English and Spanish messages contained in our working library. After these calculations, two frequency profiles (probability distributions) were obtained: one for English, the other for Spanish.

The first 25 rows of Table 1 correspond to the 25 most used symbols. After this high ranked symbols, rows in Table 1 show groups of symbols sharing ranges with the same or approximate percentage of use. In accordance with our definition of tail form this study, head-tail transition occurs at rankings 40 and 35 for English and Spanish respectively.

Joining the text messages in three sets, according to the language they are written with, we obtained an approximation of the symbol frequency profiles for the '*active*' fraction of the languages studied (see discussion). Figure 7 shows these profiles. Natural languages exhibit a wide range of ranks where the symbol frequency decays with an approximately constant slope $g$, sustaining Zipf's law for English and extending its validity to Spanish, at least up to certain range of the symbol rank dominion. Even though we included many programming languages and artificial code as if they were all part of a unique language, which they are not, artificial languages do not show a range where we can consider slope $g$ a constant, evidencing the fact that artificial languages are much smaller than natural ones. The values of exponent $g$ were calculated for the three profile tails and included in Figure 7; profile slopes are all negative but $g$ values are shown positive to be consistent with equation (8). Notice that Spanish has, among the languages studied here, the smallest tail slope, meaning the heaviest tail; an indication of the variety of words included in all the Spanish messages. At the other end of our sample, artificial languages present the fastest decaying slope and the most limited number of symbols.

| Natural languages symbol frequency | | | | | |
|---|---|---|---|---|---|
| **English. Total symbols = 23398** | | | **Spanish. Total Symbols = 33249** | | |
| **Rank** | **Word (Symbol)** | **Use [%]** | **Rank** | **Word (Symbol)** | **Use [%]** |
| 1 | the | 5.51921 | 1 | , | 5.7697 |
| 2 | , | 4.96449 | 2 | de | 5.0643 |
| 3 | . | 4.58479 | 3 | . | 3.8664 |
| 4 | of | 2.96836 | 4 | la | 3.5446 |
| 5 | and | 2.89258 | 5 | que | 3.0410 |
| 6 | to | 2.39816 | 6 | y | 2.8992 |
| 7 | a | 1.71795 | 7 | el | 2.3789 |
| 8 | in | 1.63451 | 8 | en | 2.0957 |
| 9 | that | 1.42234 | 9 | a | 1.9270 |
| 10 | i | 1.33711 | 10 | los | 1.5953 |
| 11 | is | 1.29327 | 11 | no | 1.1690 |
| 12 | it | 1.09772 | 12 | las | 0.9659 |
| 13 | we | 1.09103 | 13 | un | 0.9562 |
| 14 | not | 0.79216 | 14 | se | 0.9486 |
| 15 | " | 0.78874 | 15 | con | 0.8530 |
| 16 | for | 0.73284 | 16 | del | 0.8395 |
| 17 | he | 0.70253 | 17 | por | 0.7923 |
| 18 | have | 0.70204 | 18 | una | 0.7836 |
| 19 | was | 0.63881 | 19 | para | 0.6962 |
| 20 | be | 0.62708 | 20 | es | 0.6939 |
| 21 | this | 0.55440 | 21 | - | 0.6241 |
| 22 | as | 0.54185 | 22 | lo | 0.6229 |
| 23 | you | 0.53549 | 23 | su | 0.5637 |
| 24 | are | 0.53370 | 24 | al | 0.4811 |
| 25 | with | 0.52637 | 25 | más | 0.4503 |
| 26 | they | 0.50694 | 26 | como | 0.4330 |
| … | … | … | … | … | … |
| 58 | ***man*** | ***0.24761*** | *58* | ***pueblo*** | ***0.1435*** |
| … | … | … | *59* | ***mundo*** | ***0.1408*** |
| 62 | ***people*** | ***0.23883*** | *60* | sobre | 0.1344 |
| … | … | … | … | … | … |
| 71 | ***world*** | ***0.17423*** | *67* | ***vida*** | ***0.1256*** |
| … | | | … | … | … |
| *500…* | *indeed…* | 0.01867... | *500…* | *poeta…* | 0.01749... |
| *…8000* | *…yard* | ...0.000732 | *...7339* | *…flujo* | ...0.000843 |
| 8002 - 9920 | *adapt - vitiated* | *0.00055* | *7340-8841* | *funda…insurgimos* | 0.000843 |
| 9923 - 13505 | *actress - Zemindars* | *0.00037* | *8842-11736* | *adictos …zumbido* | 0.000632 |
| *13506 - 23398* | *Aaron-Zulu* | *0.00018* | *11737-15622* | *abastecimientos … Zelli* | 0.000419 |
| | | | *15783-33249* | *abanderado … Xavier* | 0.000209 |

Table 1. Most frequently used symbols in English and Spanish. Open-class words are shown with italic characters. Closed-class word are shown with normal characters. Top ranked open-class words are shown with italic-bold letters.

Direct measurement of differences between profile shapes is not straight forward. We converted the symbol frequency distributions into probability distributions and graph their corresponding CDF (cumulative function distribution) shown in Figure 8. As expected, artificial languages' CDF grow faster than the others; the five hundred most frequently used symbols are enough to comprise almost 90 % of all symbols included in our list of more than 13000 artificial symbols. The first 500 words cover 74 % of the 23398 English words included in our library and 70 % for the 33249-word Spanish library.

The profile heads also reflect some differences between languages. In spite of the general faster growing English's CDF as compared with Spanish, the latter's CDF is higher up to symbol ranked about 56, where the two curves cross. This Spanish faster growing CDF within the head region implies a more intensive use of the close-words group and consequently the tendency of a more structured use of this particular language.

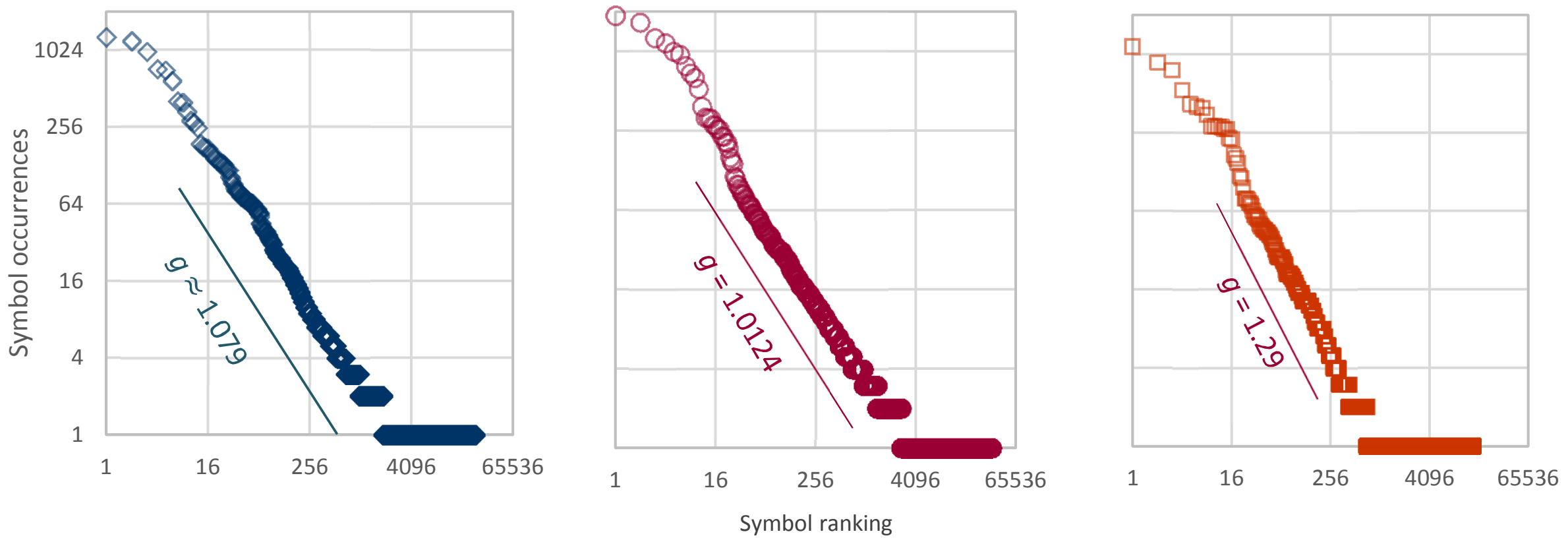

Figure 8. Ranked symbol frequency distribution for English (left), Spanish (center) and artificial languages (right).

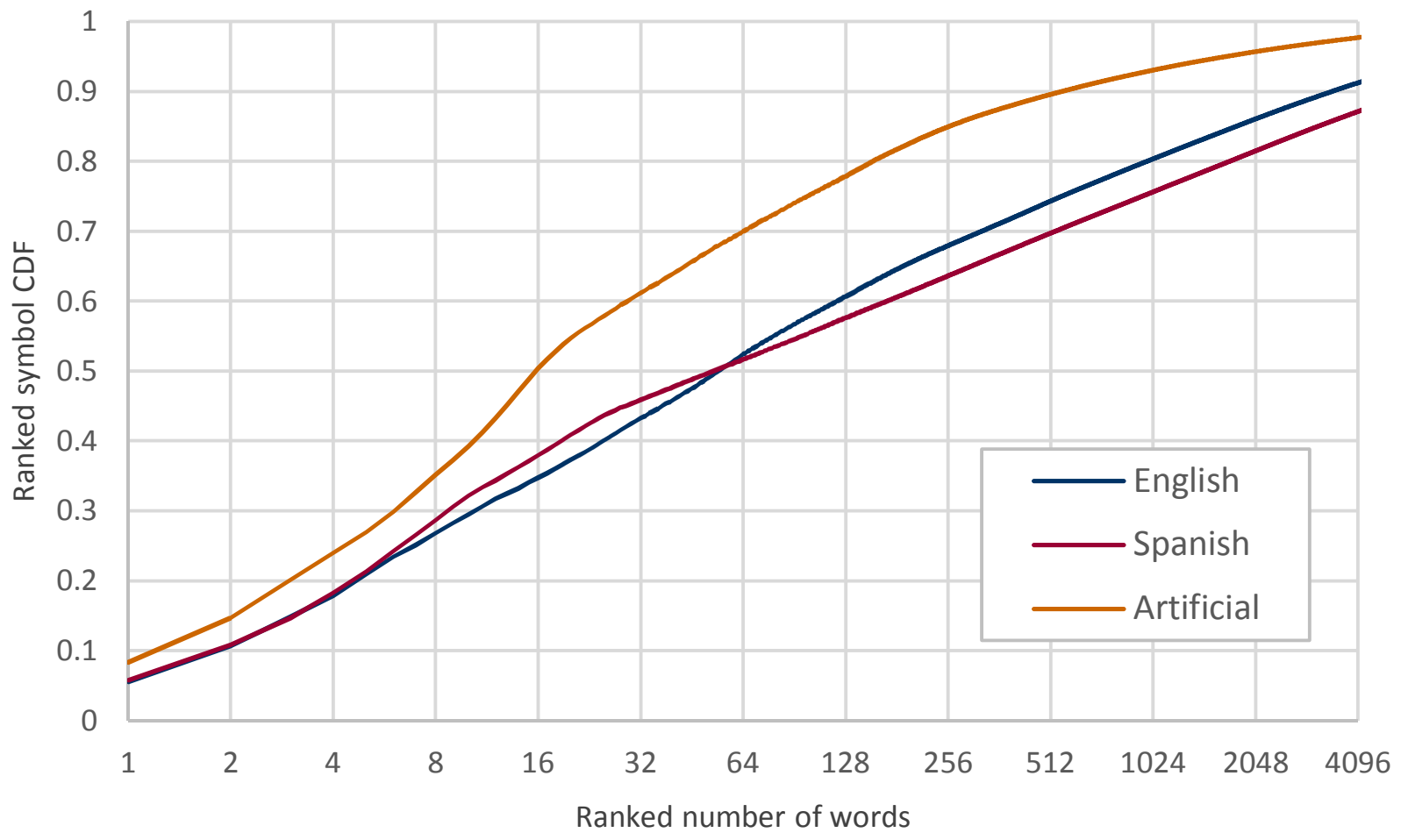

Figure 9. Cumulative distribution function (CDF) of symbols ranked by frequency. Horizontal axis is scaled to show the curves for the 4096 most frequently used words for English, Spanish and Artificial language. Note the logarithmic scale in horizontal axis.

### 3.4.1 Zipf's deviation $J_{1,D}$ for ranked distribution

We computed Zipf's deviations $J_{1,D}$ for natural and artificial languages. Figure 9 shows the result of these calculation on the plane Zipf's deviation $J_{1,D}$ vs. Length $L$. Dependence between Zipf's deviation $J_{1,D}$ and Length $L$ was evaluated with standard deviation and correlations.

We also performed two tests with Student-t distributions to compare the Zipf's deviations $J_{1,D}$ . The first tests the hypothesis of English and Spanish Zipf's distribution being the same. The second tests the hypothesis for natural and artificial languages to be the same. Results for all tests show that p-values are very small indicating that Zipfs deviation differed statistically in very significant ways between the three different languages studied. Table 2 summarizes these results.

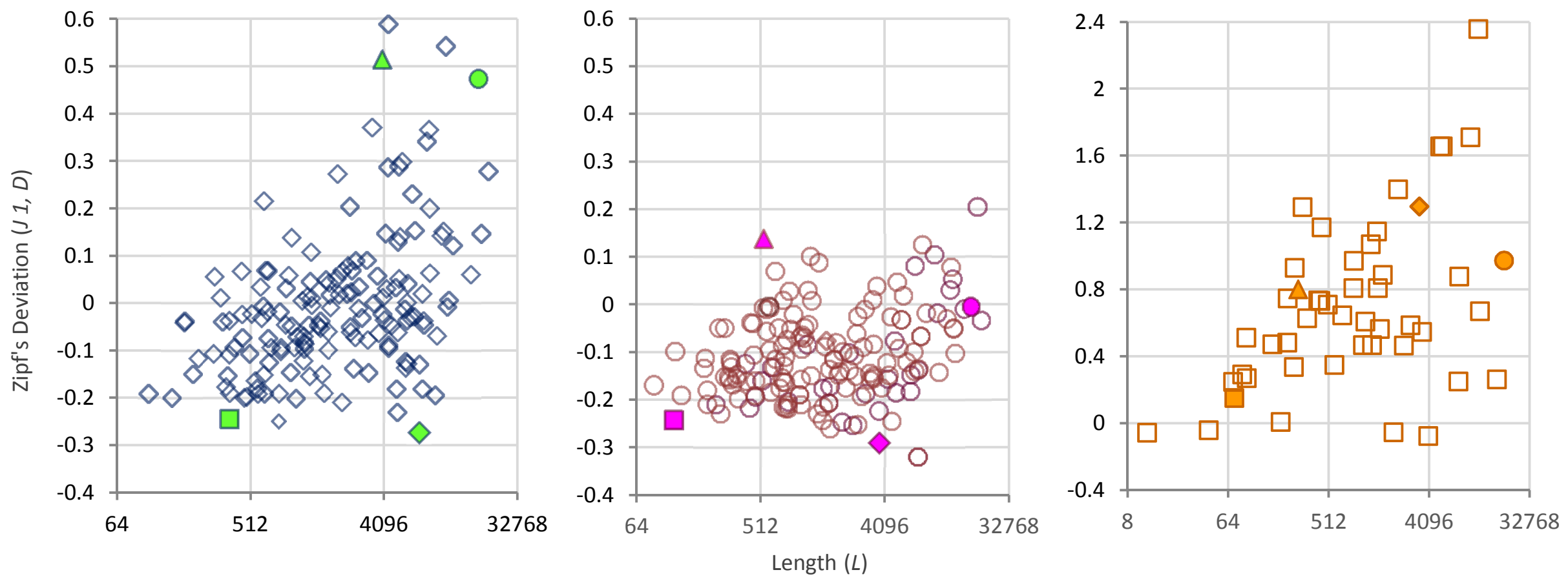

Figure 10. Zipf's deviation $J_{1,D}$ of symbol ranked frequency distributions depending on text length $L$. English (left), Spanish (center) and Software (right). Reference texts are highlighted with filled markers. English: square - *1945.BS.Eng.GabrielaMistral*, triangle - *1921.MarieCurie,* rhombus - *1950.NL.Eng.BertrandRussell,* circle - *1890.RusselConwell.* Spanish*:* square - *1936.DoloresIbarruri*, triangle - *1982.Gabriel García Márquez*, rhombus - *JoseSaramago.Valencia,* circle - *CamiloJoseCela.LaColmena.Cap1.* Artificial*:* square - *FibonacciNumbers.CSharp,* triangle - *QuickSort.CSharp,* rhombus - *Sociodynamica.Module3,* circle - *WebSite.Inmogal.php.*

**Zipfs' deviation *J 1, D* for natural and artificial languages**

| | n | J 1, D average | J 1, D Std. Dev. | Correlation J1,D : L |
|---|---|---|---|---|
| **English** | 156 | 0.0045 | 0.1719 | 0.560 |
| **Spanish** | 158 | -0.1074 | 0.0943 | 0.351 |
| **Computer Code** | 49 | 0.6944 | 0.4961 | 0.102 |

| t-test | n1 - n2 | p-value |
|---|---|---|
| **English - Spanish** | 156 - 158 | 6.58E-12 |
| **Natural - Software** | 314 - 49 | 9.47E-64 |

Table 2. Zipf's Deviation $J_{1,D}$ and its correlation with length $L$ for English, Spanish and artificial messages.

### 3.4.2 Tail Zipf's deviation $J_{\theta,D}$ for ranked tail distributions

Zipf's deviation was also inspected for the tails of the ranked frequency distributions as described in Section 2.6. This evaluation provides some further understanding of the tails shapes and relates some tendencies to other variables associated to the messages and the languages. Figure 10 shows the Zipf's deviation $J_{\theta,D}$ based on the messages tails for the three languages included in this study. The incidence of language and different group of writers over the tail of ranked frequency distributions was evaluated by performing a Student-t test which results are included in Table 3. Student-t tests to compare the distributions of the texts tail Zipf's deviations $J_{\theta,D}$ show very small p-values, indicating that tail Zipfs deviation differed statistically in very significant ways between the three different languages studied.

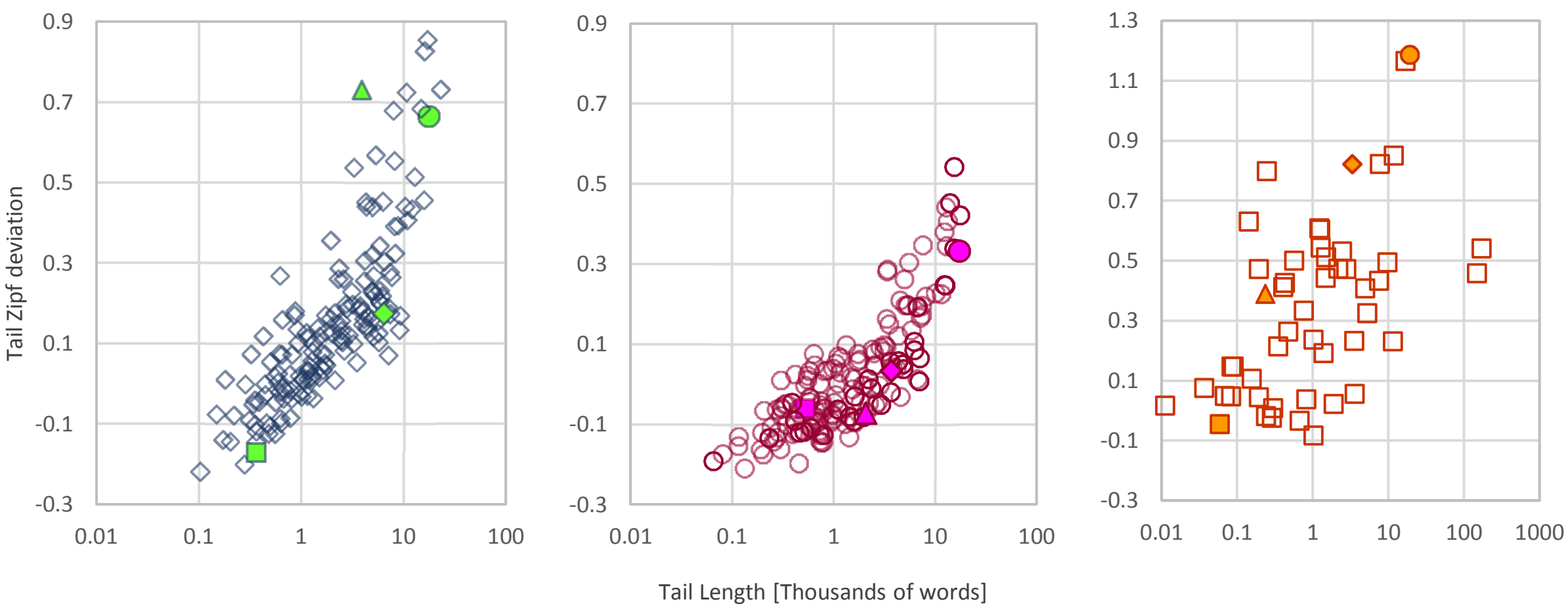

Figure 11. Tail Zipf's deviation $J_{\theta,D}$ for symbol ranked frequency distributions vs. text tail length $L$. English (left), Spanish (center) and Software (right). Reference texts are highlighted with filled markers. English: square - *1945.BS.Eng.GabrielaMistral*, triangle - *1921.MarieCurie,* rhombus - *1950.NL.Eng.BertrandRussell,* circle - *1890.RusselConwell.* Spanish*:* square - *1936.DoloresIbarruri*, triangle - *1982.Gabriel García Márquez*, rhombus - *JoseSaramago.Valencia,* circle - *CamiloJoseCela.LaColmena.Cap1.* Artificial*:* square - *FibonacciNumbers.CSharp,* triangle - *QuickSort.CSharp,* rhombus - *Sociodynamica.Module3,* circle - *WebSite.Inmogal.php.*

**Tail Zipf's deviation $J_{\theta,D}$ for natural and artificial languages**

| | n | $J_{\theta,D}$ average | $J_{\theta,D}$ Std. Dev. | *Correlation* $J_\theta$ : $L_\theta$ |
|---|---|---|---|---|
| **English** | 156 | 0.1502 | 0.2108 | 0.809 |
| **Spanish** | 158 | 0.0235 | 0.1493 | 0.856 |
| **Computer Code** | 49 | 0.3528 | 0.3062 | 0.640 |
| **t-test** | **n1 - n2** | **p-value** | | |
| **English - Spanish** | 156 - 158 | 2.34E-09 | | |
| **Natural - Software** | 314 - 49 | 2.79E-15 | | |

Table 3. Tail Zipf's deviation $J_{\theta,D}$ and its correlation with message tail length $L_\theta$ for English, Spanish and artificial messages.

# 4 Discussions

## 4.1 Diversity for Natural and Artificial Languages

Setting a precise number for the total number of words of a natural language is impossible, as words appear and disappear constantly. However it has been estimated that English contains more words than Spanish [19] [20] [21]. Living languages evolve over time and structural differences make it difficult to compare figures of language size measure. Nevertheless the numbers of lemmas in dictionaries provide us a reference to compare language sizes. The dictionary of the Real Academia Española contains 87.718 Spanish lemmas [22] while the Oxford English dictionary includes about 600.000 words [23]. Despite the larger size of English dictionaries, Spanish texts showed higher and less dispersed symbol diversity than English.

The higher word diversity of Spanish may thus be due to factors such as syntactical rules or grammar which affect both languages differently. Verb tenses and conjugations, for example, are all considered as one word when included in a dictionary, but each of them was recognized as a different symbol here.

For Spanish, most articles, pronouns and subject genres vary from masculine to feminine while for English this only happens for particular cases like *his*/*her*. These grammar characteristics may increase the number of different symbols used in any Spanish texts, but considering the relative size of closed and open word groups, this effect should be marginal with regard to general text symbol diversity. On the other hand, verbs, which belong to the open group of words, have more tenses and conjugations for Spanish and therefore increase Spanish word diversity in ways not accounted for in dictionaries. Grammar is then one feature that explains greater Spanish word diversity compared to English.

These differences might explain only parts of the results shown here. A wider use of words in Spanish, compared to English, despite a larger number of words in English dictionaries, cannot be excluded.

## 4.2 Entropy for Natural and Artificial Languages

There is no qualitative difference for this property between English and Spanish, perhaps a consequence of the similar structure and functionality both natural languages share. Nevertheless entropy appears slightly higher for messages expressed in English than for those in Spanish; being English a larger language in terms of words, this result might be explained as consequence of a more elaborated grammar in Spanish allowing for lower entropy levels.

Natural languages have developed to express concepts and complex ideas. Natural languages can express many different types of messages such as information, persuasion, inspiration, instruction, distraction and joy. Artificial languages, in contrast, are designed to give precise instructions; they are more formal than natural ones [24] as they must convey precise and unequivocal information to machines. Artificial languages are represented by computer programs; collections of instructions having extensive number of symbols and commands. The number of symbols that an artificial language usually contains is very small when compared to natural ones. Essentially there are neither synonyms nor pronouns. Connecting and auxiliary words like prepositions and articles are limited to conditional and logical expressions. Adjectives are replaced by numeric variables which may quantify some aspects modeled. With these limitations, computer languages have little room for style compared to natural languages. Computer code is valued for its effectiveness rather than its beauty. The limited structure to form sentences in artificial languages leads to a relatively flatter frequency distribution and therefore higher entropy levels.

Since emergence is defined as equivalent to Shannon's information (entropy), the higher emergence for artificial languages implies that less symbols are used to produce more meaning. In other words, there is less redundancy in artificial than in natural languages. Redundancy can lead to robustness [25], which is desirable in natural languages where communication may be noisy. However, artificial languages are created for formal, deterministic compliers or interpreters, so there is no pressure to develop robustness.

Self-organization, as opposed to emergence, is higher in artificial than in natural languages. This is because of the same reason explained above: artificial languages require more structure to be more precise, which fulfills their purpose. Natural languages are less organized because they require flexibility and adaptability for their purpose, which includes the ability of having different words with the same meaning (synonymy) and words with different meanings (polysemy).

For the same specific diversity $\boldsymbol{d}$, complexity is higher for natural languages (Figure 5). However, for the same length $\boldsymbol{L}$, complexity is higher for artificial languages, as emergence dominates the properties of all languages ($\boldsymbol{e} > \mathbf{0.5}$) (Figure 6). Artificial languages are slightly more regular, but all languages have a relatively high entropy and thus emergence.

### 4.3 Symbol Frequency Distributions

Intuition may suggest that the symbol frequency profile of a symbol limited language will decay faster than a richer language in terms of number of available symbols. Figures 8 illustrates how, for the natural languages considered here, the points of each message rank distribution profile lay close to a straight line connecting the first with the last ranked word. This indicates that $g$ values for natural languages are approximately constant over the range of symbol ranking. For artificial texts, on the contrary, symbol-frequency vs. symbol-ranking does not show a constant decay value. The slope of the graph is low for most used symbols and increases its decay rate as the symbols considered approach the least used ones, giving the rank symbol profile of artificial language the concave downward shape characteristic of an approximation to the cut-off region [26]. This increasing slope $g$ that artificial messages exhibit over ranges of the ranking dominion indicate these languages are close to the physical limit of their total number of symbols. For natural languages $g$ values are not only lower but also closer to a constant, denoting that natural language profiles are within the scale-free region and therefore far from the physical limit [26] imposed by the number of symbols they are constituted with. Natural languages are significantly larger than the artificial languages all together.

There is a qualitative difference of the symbol frequency distributions for natural and artificial languages; texts written in natural languages correlate with a power law distribution for all the Symbol Ranking ranges while artificial texts show an increasing decay slope for ranges of least used symbols. This difference may be related to the fact that for natural languages any message uses only a tiny fraction of the whole set of words of the language, while any reasonable long computer code will use a large fraction of the whole set of symbols available in the computer language.

The most conspicuous difference between natural and artificial languages was revealed using ZIpf's deviation $J_{1,D}$. Statistical analysis revealed highly significant differences between natural and artificial languages in this variable. Tail Zipf's deviation $J_{\theta,D}$ , confirmed these differences, focusing only on the tails of these distributions. No loss of information was evidenced when focusing our analysis only on the tails, compared with analysis using the complete frequency profile of the ZIpf's deviation $J_{1,D}$.

Another interesting aspect of this list of symbols is where the words of open and close classes lay according to their frequency of use; close and open word classes are also known as core and non-core word types. As Andrew Moore explained [27], English grew by adding new words to its open-word class consisting of nouns, verbs and qualifiers, (adjectives and adverbs). The close-word class contains determiners, pronouns, prepositions and conjunctions; words that establish functionality and language structure.  The dynamic process of word creation and the 'flow' of words from one class to the other have been recently modeled [28]. Changes over time are slow, thus for our purpose of this study, we considered the open and close classes as invariant groups. Being the open-class the sustained faster growing type of words of natural languages, it is reasonable to expect the open words class to be much larger than the group of closed words. The smaller size of the closed-word class and the highly restricted character of its components (most of them do not even have synonyms), explain the high frequency of their use and their tendency to be placed near the top of the ranked list shown in Table 1, letting the open-class words to sink down to lower ranked positions of the list. There are formal indications of this tendency of close words to group near the top of frequency ranked list in a study by Montemurro et al. [29], where pronouns are presented as the most frequently used word-function in Shakespeare's Hamlet.

Besides being necessary to understand the structure of English and Spanish, the classification of words as members of the open and closed groups is important because analyzing the ranking among the open-class words may lead to some practical uses as the recognition of message subject or theme. The highest ranked open-class words are represented using italic-bold characters. For the messages included in this study, the most used open-class words were '*man'*, '*people' and 'world'* for English, and '*pueblo'*, '*mundo'* and '*vida'* for Spanish; all of them are terms with strong connection to government, religion, and human rights as the main theme treated by the majority of the messages.

# 5 Conclusions

Diversity is higher for Spanish messages than for English ones, suggesting that there is influence of cultural constraints over message diversity. Being more restricted to very specific uses and less dependent on writing style, artificial languages showed a considerably lower diversity than natural languages.

Entropy measures for natural languages are higher than those for artificial. The larger symbolic diversity for natural languages dominates the resulting text entropies, leaving frequency profiles to a more subtle influence. When comparing English and Spanish however, symbolic diversities are closer to each other while entropy differences become relevant. Future work could include sets of legal, clinical or technical documents. Since these seem to be more specific, they should have properties in between the natural and artificial sets studied here.

We have shown that important differences among languages become evident by experimentally measuring symbolic diversity, emergence and complexity in collections of texts. The differences detected are the result of the combination of the current status of their respective evolution as well as cultural aspects that affect the style of communicating and writing.

These differences among languages are evidenced measuring symbolic diversity, emergence and complexity in collections of texts. Yet the most reliable measure was the symbolic diversity. Applying this procedure over the basis of a 'grammar scale complexity' would provide a deeper sense of languages nature and behavior.

We believe that the present study showed that complexity analysis can add to our understanding of features of natural languages. For example, automatic devises to differentiate text written by computers from text produced by real persons might be feasible using this knowledge. Yet our study also revealed that Complexity Science is in a very incipient state regarding its capacity to extract meaning from the analysis of texts. Much interesting work lies ahead.

# Appendix A

Artificial texts:

http://www.gfebres.com/F0IndexFrame/F132Body/F132BodyPublications/NatArtifLangs/Whole/Artificial.Properties.htm

## Artificial language texts properties

***L*** Text Length
***D*** Diversity
***d*** Specific diversity [0-1]
***h*** Entropy [0-1]
***g*** Zipf's exponent
$J_{1,D}$ Zipf's diviation
$J_{\vartheta,D}$ Tail Zipf's diviation

| Text Name | *L* | *D* | *d* | *h* | *g* | $J_{1,D}$ | $J_{\vartheta,D}$ |
|---|---|---|---|---|---|---|---|
| FibonacciNumbers.CSharp | 62 | 27 | 0.435 | 0.921 | 0.788 | 0.100 | -0.045 |
| Math.Mime2d.MathLab | 11376 | 120 | 0.011 | 0.681 | 1.599 | 2.334 | 0.249 |
| Levenberg.MathLab | 567 | 99 | 0.175 | 0.823 | 0.993 | 0.376 | 0.502 |
| IsPrime.C | 158 | 56 | 0.354 | 0.902 | 0.823 | 0.146 | 0.108 |
| InsertAfterBefore.CSharp | 141 | 37 | 0.262 | 0.935 | 0.738 | 0.615 | 0.632 |
| MathLab.Fr.MathLab | 1707 | 207 | 0.121 | 0.788 | 1.041 | 0.561 | 0.583 |
| MathLab.pplane8.MathLab | 68788 | 2157 | 0.031 | 0.586 | 1.252 | 0.435 | 1.751 |
| MatrixLUDecomp.CSharp | 416 | 52 | 0.125 | 0.856 | 1.007 | 1.011 | 0.413 |
| MatrixFuncts.CSharp | 8069 | 194 | 0.024 | 0.663 | 1.487 | 0.704 | 0.565 |
| MathLab.Taller.MathLab | 2162 | 122 | 0.056 | 0.740 | 1.190 | 1.383 | 0.472 |
| MathLab.programa2.MathLab | 9324 | 254 | 0.027 | 0.692 | 1.345 | 1.555 | 0.476 |
| HeapSort.Java | 314 | 59 | 0.188 | 0.857 | 0.956 | 0.534 | 0.010 |
| HeapSort.CSharp | 247 | 46 | 0.186 | 0.900 | 0.894 | 0.721 | 0.388 |
| HanoiTowers.Java | 484 | 92 | 0.190 | 0.847 | 0.892 | 0.624 | 0.266 |
| CopyFolderNContent.CSharp | 195 | 49 | 0.251 | 0.908 | 0.800 | 0.488 | 0.473 |
| ChainedScatterTable.CSharp | 201 | 46 | 0.229 | 0.890 | 0.842 | 0.570 | 0.046 |
| BoolFunctWithMultiplexerLogic.C | 1030 | 163 | 0.158 | 0.796 | 1.000 | 0.479 | 0.238 |
| BlowfishEncryption.C | 3574 | 669 | 0.187 | 0.706 | 1.254 | -0.181 | 0.058 |
| ExtendedEuclidean.C | 86 | 24 | 0.279 | 0.902 | 0.846 | 0.443 | 0.149 |
| GameOfLife.C | 247 | 46 | 0.186 | 0.893 | 0.864 | 0.882 | 0.799 |
| FTPFunctions.CSharp | 11505 | 312 | 0.027 | 0.593 | 1.467 | 0.286 | 1.184 |
| FiniteElements.MathLab | 2748 | 295 | 0.107 | 0.731 | 1.079 | 0.551 | 0.467 |
| Factorial.CSharp | 36 | 21 | 0.583 | 0.965 | 0.578 | 0.063 | 0.078 |
| MatrixLUDecomp.Phyton | 298 | 40 | 0.134 | 0.882 | 0.956 | 1.250 | -0.023 |
| MetaWords.FormsAnsClasses.CSharp | 1279 | 144 | 0.113 | 0.828 | 0.963 | 1.065 | 0.544 |
| Sociodynamica.Module1 | 9617 | 290 | 0.030 | 0.666 | 1.271 | 1.705 | 0.496 |
| Sociodynamica.Forms | 2428 | 297 | 0.122 | 0.759 | 1.075 | 0.461 | 0.532 |
| SnakeGame.C | 1515 | 157 | 0.104 | 0.803 | 1.061 | 0.816 | 0.512 |
| QuickSort.CSharp | 364 | 56 | 0.154 | 0.896 | 0.882 | 0.934 | 0.216 |
| Sociodynamica.Module2 | 7672 | 428 | 0.056 | 0.706 | 1.176 | 0.873 | 0.824 |
| Sociodynamica.Module3 | 3363 | 223 | 0.066 | 0.770 | 1.086 | 1.288 | 0.822 |
| Sumation.CSharp | 71 | 25 | 0.352 | 0.895 | 0.850 | 0.208 | 0.051 |
| WebSite.TiempoReal.Html | 7495 | 565 | 0.075 | 0.586 | 1.264 | 0.250 | 0.434 |
| WebSite.RistEuropa.Html | 11713 | 503 | 0.043 | 0.595 | 1.321 | 0.668 | 0.852 |
| WebSite.Inmogal.php | 19299 | 647 | 0.034 | 0.632 | 1.261 | 0.966 | 1.185 |
| ViscomSoft.ScannerActivex.CSharp | 11275 | 623 | 0.055 | 0.534 | 1.405 | -0.084 | 0.471 |
| QuadraticPrograming.CSharp | 433 | 72 | 0.166 | 0.848 | 0.927 | 0.713 | 0.427 |
| Polinom.CSharp | 86 | 33 | 0.384 | 0.916 | 0.760 | 0.228 | 0.049 |
| PermutationAlgorithm.Java | 1227 | 96 | 0.078 | 0.775 | 1.149 | 1.032 | 0.610 |
| NetPlexMainForm.CSharp | 59496 | 1218 | 0.020 | 0.531 | 1.432 | 0.232 | 1.449 |
| NetPlex.Forms.CSharp | 66281 | 1482 | 0.022 | 0.644 | 1.226 | 1.492 | 1.449 |
| NetPlex.Classes.CSharp | 18662 | 652 | 0.035 | 0.689 | 1.139 | 1.880 | 1.416 |
| ModularInverse.C | 91 | 31 | 0.341 | 0.902 | 0.865 | 0.248 | 0.148 |
| MetaWordsMainForm.CSharp | 65492 | 1081 | 0.017 | 0.531 | 1.462 | 0.279 | 1.478 |
| PartDifEqtnsHeatEq.MathLab | 677 | 104 | 0.154 | 0.774 | 1.053 | 0.613 | -0.041 |
| PartDifEqtnsLaplaceEq.MathLab | 825 | 96 | 0.116 | 0.780 | 1.131 | 0.741 | 0.030 |
| PermutationAlgorithm.Csharp | 777 | 88 | 0.113 | 0.841 | 1.015 | 1.041 | 0.334 |
| PartDifEqtnsWaveEqtn.MathLab | 249 | 67 | 0.269 | 0.855 | 0.897 | 0.336 | -0.017 |
| mail.log.2 | 147418 | 3195 | 0.022 | 0.560 | 1.297 | 1.297 | 0.460 |
| Apache.Access.log | 168355 | 2870 | 0.017 | 0.534 | 1.464 | 0.445 | 0.542 |

English texts (1/3):
http://www.gfebres.com/F0IndexFrame/F132Body/F132BodyPublications/NatArtifLangs/Whole/English.Properties.htm

## English texts properties

| | | |
|---|---|---|
| ***L*** Text Length | ***d*** Specific diversity [0-1] | $J_{1,D}$ Zipf's diviation |
| ***D*** Diversity | ***h*** Entropy [0-1] | $J_{\vartheta,D}$ Tail Zipf's diviation |
| | ***g*** Zipf's exponent | |

| Text Name | *L* | *D* | *d* | *h* | *g* | $J_{1,D}$ | $J_{\vartheta,D}$ |
|---|---|---|---|---|---|---|---|
| 1381.JohnBall | 227 | 117 | 0.5154 | 0.9143 | 0.7680 | -0.1156 | -0.0788 |
| 1588.QueenElizabethI | 359 | 156 | 0.4345 | 0.8786 | 0.9376 | -0.1528 | -0.0414 |
| 1601.Hamlet | 150 | 97 | 0.6467 | 0.9501 | 0.7449 | -0.1987 | -0.0760 |
| 1601.QueenElizabethI | 1140 | 388 | 0.3404 | 0.8647 | 0.8467 | 0.0056 | 0.1264 |
| 1606.LancelotAndrewes | 9285 | 1540 | 0.1659 | 0.7374 | 1.0936 | -0.0690 | 0.1691 |
| 1814.NapoleonBonaparte | 182 | 95 | 0.5220 | 0.9202 | 0.7380 | -0.0406 | -0.1394 |
| 1833.ThomasBabington | 15668 | 2647 | 0.1689 | 0.7460 | 0.9988 | 0.0602 | 0.4564 |
| 1849.LucretiaMott | 7575 | 1720 | 0.2271 | 0.7705 | 0.9759 | -0.0457 | 0.2653 |
| 1851.ErnestineLRose | 8301 | 1630 | 0.1964 | 0.7643 | 0.9851 | 0.0630 | 0.3239 |
| 1851.SojournerTruth | 436 | 180 | 0.4128 | 0.9185 | 0.7197 | 0.0651 | 0.1175 |
| 1861.AbrahamLincoln | 4007 | 1018 | 0.2541 | 0.8077 | 0.9268 | 0.0116 | 0.1815 |
| 1863.AbrahamLincoln | 292 | 143 | 0.4897 | 0.9270 | 0.6524 | 0.0557 | -0.0018 |
| 1867.ElizabethCadyStanton | 5847 | 1481 | 0.2533 | 0.7846 | 0.9869 | -0.1277 | 0.2027 |
| 1873.SusanBAnthony | 626 | 255 | 0.4073 | 0.8617 | 0.9043 | -0.1505 | -0.0215 |
| 1877.ChiefJoseph | 183 | 92 | 0.5027 | 0.9257 | 0.7401 | -0.0375 | 0.0102 |
| 1890.RusselConwell | 17766 | 2207 | 0.1242 | 0.7483 | 0.9861 | 0.4742 | 0.6646 |
| 1892.FrancesEWHarper | 4395 | 1244 | 0.2830 | 0.8053 | 0.9417 | -0.0930 | 0.1430 |
| 1901.MarkTwain | 660 | 255 | 0.3864 | 0.8889 | 0.7560 | 0.0641 | 0.0714 |
| 1903.BS.Eng.BjornstjerneBjornson | 1651 | 573 | 0.3471 | 0.8496 | 0.8244 | 0.0487 | 0.0336 |
| 1906.MaryChurch | 1558 | 585 | 0.3755 | 0.8524 | 0.8963 | -0.1499 | 0.0423 |
| 1909.BS.SelmaLagerlof | 2296 | 626 | 0.2726 | 0.8247 | 0.9300 | 0.0186 | 0.1479 |
| 1915.AnnaHoward | 10633 | 1425 | 0.1340 | 0.7747 | 0.9506 | 0.5420 | 0.7245 |
| 1916.CarrieChapman | 6120 | 1542 | 0.2520 | 0.7943 | 0.9360 | -0.0303 | 0.2165 |
| 1916.HellenKeller | 2557 | 854 | 0.3340 | 0.8294 | 0.9202 | -0.1370 | 0.1029 |
| 1918.WoodrowWilson | 2753 | 769 | 0.2793 | 0.8177 | 0.9053 | -0.0112 | 0.1972 |
| 1920.CrystalEastman | 2131 | 669 | 0.3139 | 0.8482 | 0.8341 | 0.0675 | 0.1753 |
| 1921.MarieCurie | 921 | 307 | 0.3333 | 0.8642 | 0.8356 | 0.0582 | 0.1353 |
| 1923.BS.Eng.WilliamButlerYeats | 320 | 167 | 0.5219 | 0.9197 | 0.6626 | 0.0119 | -0.0906 |
| 1923.JamesMonroe | 1178 | 417 | 0.3540 | 0.8493 | 0.8830 | -0.0816 | 0.0324 |
| 1923.NL.Eng.WilliamButlerYeats | 4257 | 1127 | 0.2647 | 0.8194 | 0.9079 | 0.0306 | 0.3058 |
| 1925.MaryReynolds | 4300 | 852 | 0.1981 | 0.8022 | 0.9134 | 0.2870 | 0.4510 |
| 1930.NL.Eng.SinclairLewis | 5707 | 1609 | 0.2819 | 0.7986 | 0.9534 | -0.1360 | 0.1034 |
| 1932.MargaretSanger | 1162 | 399 | 0.3434 | 0.8468 | 0.9113 | -0.1212 | 0.1130 |
| 1936.EleanorRoosvelt | 1966 | 457 | 0.2325 | 0.8301 | 0.8796 | 0.2744 | 0.3562 |
| 1936.KingEdwardVIII | 596 | 243 | 0.4077 | 0.8747 | 0.7740 | 0.0342 | 0.0039 |
| 1936.NL.Eng.EugeneOneill | 1177 | 407 | 0.3458 | 0.8381 | 0.8972 | -0.0955 | 0.0109 |
| 1938.BS.PearlBuck | 520 | 197 | 0.3788 | 0.8935 | 0.8255 | -0.0226 | 0.0538 |
| 1938.NL.PearlBuck | 10270 | 1825 | 0.1777 | 0.7666 | 0.9751 | 0.1511 | 0.4402 |
| 1940.05.WinstonChurchill | 703 | 292 | 0.4154 | 0.8730 | 0.8546 | -0.0931 | -0.0164 |
| 1940.06.A.WinstonChurchill | 3762 | 1067 | 0.2836 | 0.8228 | 0.9142 | -0.0642 | 0.1929 |
| 1940.06.B.WinstonChurchill | 4899 | 1189 | 0.2427 | 0.8022 | 0.9318 | 0.0320 | 0.2297 |
| 1941.AdolfHitler | 10901 | 2228 | 0.2044 | 0.7691 | 0.9748 | -0.0076 | 0.4059 |
| 1941.FranklinDRoosvelt | 574 | 261 | 0.4547 | 0.8806 | 0.9119 | -0.1906 | -0.0467 |
| 1941.HaroldIckes | 2448 | 720 | 0.2941 | 0.8221 | 0.8930 | 0.0248 | 0.1185 |
| 1942.MahatmaGandhi | 1234 | 428 | 0.3468 | 0.8554 | 0.8496 | 0.0138 | 0.0437 |
| 1944.DwightEisenhower | 208 | 120 | 0.5769 | 0.9248 | 0.7615 | -0.1486 | -0.1436 |
| 1944.GeorgePatton | 873 | 313 | 0.3585 | 0.8861 | 0.8021 | 0.0323 | 0.1694 |
| 1945.BS.Eng.GabrielaMistral | 370 | 196 | 0.5297 | 0.8833 | 0.9373 | -0.2446 | -0.1705 |

English texts (cont. 2/3):

| | | | | | | | |
|---|---|---|---|---|---|---|---|
| 1946.WinstonChurchill | 1285 | 498 | 0.3875 | 0.8496 | 0.9141 | -0.1538 | 0.0101 |
| 1947.GeorgeCMarshall | 1606 | 582 | 0.3624 | 0.8422 | 0.9275 | -0.1892 | 0.0749 |
| 1947.HarryTruman | 2445 | 716 | 0.2928 | 0.8218 | 0.9296 | -0.0612 | 0.1564 |
| 1948.BS.Eng.ThomasEliot | 1467 | 504 | 0.3436 | 0.8448 | 0.8791 | -0.0470 | 0.1376 |
| 1949.BS.Eng.WilliamFaulkner | 622 | 248 | 0.3987 | 0.8838 | 0.7906 | -0.0064 | 0.0733 |
| 1950.MargaretChase | 1717 | 561 | 0.3267 | 0.8450 | 0.8948 | -0.0602 | 0.0507 |
| 1950.NL.Eng.BertrandRussell | 6476 | 1590 | 0.2455 | 0.7893 | 0.9457 | -0.0280 | 0.1745 |
| 1953.DwightEisenhower | 2906 | 830 | 0.2856 | 0.8108 | 0.9130 | -0.0310 | 0.1212 |
| 1953.NelsonMandela | 4967 | 1433 | 0.2885 | 0.8010 | 0.9572 | -0.1801 | 0.1607 |
| 1954.BS.Eng.ErnestHemingway | 367 | 183 | 0.4986 | 0.9195 | 0.7180 | -0.0366 | -0.1006 |
| 1957.MartinLutherKing | 7952 | 1261 | 0.1586 | 0.7797 | 0.9508 | 0.3431 | 0.6792 |
| 1959.RichardFeynman | 8135 | 1300 | 0.1598 | 0.7855 | 0.9460 | 0.3665 | 0.5540 |
| 1961.01.JohnFKennedy | 1519 | 529 | 0.3483 | 0.8523 | 0.8439 | -0.0371 | 0.1181 |
| 1961.04.JohnFKennedy | 1715 | 605 | 0.3528 | 0.8454 | 0.8870 | -0.0980 | 0.0729 |
| 1961.05.JohnFKennedy | 6584 | 1535 | 0.2331 | 0.7991 | 0.8770 | 0.1528 | 0.3033 |
| 1961.11.JohnFKennedy | 680 | 316 | 0.4647 | 0.8922 | 0.8045 | -0.1246 | -0.0369 |
| 1962.09.JohnFKennedy | 2428 | 751 | 0.3093 | 0.8272 | 0.8962 | -0.0497 | 0.1076 |
| 1962.10.JohnFKennedy | 2772 | 811 | 0.2926 | 0.8287 | 0.8727 | 0.0344 | 0.1864 |
| 1962.12.MalcomX | 17199 | 1640 | 0.0954 | 0.7573 | 1.0099 | 0.7080 | 0.8545 |
| 1962.BS.Eng.JohnSteinbeck | 952 | 385 | 0.4044 | 0.8589 | 0.8830 | -0.1454 | -0.0257 |
| 1963.06.10.JohnFKennedy | 3680 | 1019 | 0.2769 | 0.8149 | 0.8815 | 0.0568 | 0.1859 |
| 1963.06.26.JohnFKennedy | 662 | 237 | 0.3580 | 0.8752 | 0.8416 | -0.0101 | 0.1583 |
| 1963.09.20.JohnFKennedy | 3986 | 1089 | 0.2732 | 0.8043 | 0.9247 | -0.0233 | 0.1468 |
| 1963.MartinLutherKing | 1731 | 527 | 0.3044 | 0.8366 | 0.8858 | 0.0428 | 0.1703 |
| 1964.04.MalcomX | 3288 | 660 | 0.2007 | 0.8198 | 0.8880 | 0.3468 | 0.5379 |
| 1964.05.LyndonBJohnson | 1168 | 430 | 0.3682 | 0.8484 | 0.8902 | -0.0695 | -0.0292 |
| 1964.LadybirdJohnson | 818 | 353 | 0.4315 | 0.8762 | 0.8278 | -0.1008 | -0.0384 |
| 1964.MartinLutherKing | 1266 | 498 | 0.3934 | 0.8616 | 0.8238 | -0.0458 | 0.0172 |
| 1964.NelsonMandela | 11929 | 2152 | 0.1804 | 0.7667 | 0.9602 | 0.1212 | 0.4337 |
| 1965.03.LyndonBJohnson | 4166 | 975 | 0.2340 | 0.8052 | 0.9006 | 0.1485 | 0.2551 |
| 1965.04.LyndonBJohnson | 1286 | 419 | 0.3258 | 0.8486 | 0.8581 | 0.0454 | 0.0569 |
| 1967.BS.Eng.MiguelAngelAsturias | 1039 | 435 | 0.4187 | 0.8489 | 0.8953 | -0.2004 | 0.0091 |
| 1967.MartinLutherKing | 7360 | 1743 | 0.2368 | 0.7939 | 0.9337 | 0.0198 | 0.2773 |
| 1967.NL.Eng.MiguelAngelAsturias | 5026 | 1479 | 0.2943 | 0.7899 | 0.9974 | -0.2295 | 0.1170 |
| 1968.MartinLutherKing | 5022 | 986 | 0.1963 | 0.7928 | 0.9192 | 0.2791 | 0.4393 |
| 1968.RobertFKennedy | 627 | 197 | 0.3142 | 0.8978 | 0.7639 | 0.2170 | 0.2682 |
| 1969.IndiraGhandi | 1058 | 408 | 0.3856 | 0.8674 | 0.8591 | -0.0709 | 0.0118 |
| 1969.RichardNixon | 5056 | 1105 | 0.2186 | 0.8048 | 0.9198 | 0.1308 | 0.3209 |
| 1969.ShirleyChisholm | 966 | 382 | 0.3954 | 0.8671 | 0.8312 | -0.0465 | 0.0192 |
| 1971.BS.Eng.PabloNeruda | 503 | 209 | 0.4155 | 0.8701 | 0.8653 | -0.1067 | -0.1051 |
| 1971.NL.Eng.PabloNeruda | 4114 | 1150 | 0.2795 | 0.8115 | 0.9101 | -0.0349 | 0.1604 |
| 1972.JaneFonda | 792 | 340 | 0.4293 | 0.8741 | 0.8661 | -0.1875 | -0.0853 |
| 1972.RichardNixon | 5362 | 920 | 0.1716 | 0.7938 | 0.9442 | 0.3032 | 0.5676 |
| 1974.RichardNixon | 1959 | 536 | 0.2736 | 0.8331 | 0.8944 | 0.0518 | 0.1618 |
| 1976.BS.Eng.SaulBellow | 395 | 199 | 0.5038 | 0.9118 | 0.7613 | -0.0981 | -0.0452 |
| 1976.NL.Eng.SaulBellow | 5625 | 1499 | 0.2665 | 0.7990 | 0.9128 | -0.0142 | 0.1551 |
| 1977.NL.Eng.VicenteAleixandre | 2618 | 845 | 0.3228 | 0.8267 | 0.9052 | -0.0723 | 0.0825 |
| 1979.MargaretThatcher | 3217 | 1002 | 0.3115 | 0.8204 | 0.9330 | -0.1461 | 0.0988 |
| 1979.MotherTeresa | 4349 | 652 | 0.1499 | 0.8055 | 0.9244 | 0.5909 | 0.4403 |
| 1981.RonaldReagan | 1175 | 448 | 0.3813 | 0.8553 | 0.8701 | -0.0838 | 0.0317 |
| 1982.NL.Eng.GabrielGarciaMarquez | 2132 | 833 | 0.3907 | 0.8401 | 0.9209 | -0.2080 | 0.0088 |
| 1982.RonaldReagan | 5037 | 1392 | 0.2764 | 0.8062 | 0.9200 | -0.0617 | 0.2275 |
| 1983.BS.Eng.WilliamGolding | 369 | 201 | 0.5447 | 0.9131 | 0.8124 | -0.1896 | -0.1445 |
| 1983.NL.Eng.WilliamGolding | 5140 | 1375 | 0.2675 | 0.8124 | 0.8904 | 0.0502 | 0.2202 |
| 1983.RonaldReagan | 5160 | 1257 | 0.2436 | 0.8137 | 0.8787 | 0.1348 | 0.2692 |
| 1986.BS.Eng.WoleSoyinka | 482 | 245 | 0.5083 | 0.8925 | 0.8611 | -0.1972 | -0.1240 |
| 1986.NL.Eng.WoleSoyinka | 9033 | 2530 | 0.2801 | 0.7783 | 0.9740 | -0.1930 | 0.1335 |

English texts (cont. 3/3):

| | | | | | | | |
|---|---|---|---|---|---|---|---|
| 1986.RonaldReagan | 784 | 305 | 0.3890 | 0.8603 | 0.8460 | -0.0231 | 0.0221 |
| 1987.RonaldReagan | 3160 | 935 | 0.2959 | 0.8218 | 0.9151 | -0.0794 | 0.1962 |
| 1988.AnnRichards | 3109 | 863 | 0.2776 | 0.8265 | 0.8720 | 0.0951 | 0.2281 |
| 1989.BS.Eng.CamioJoseCela | 464 | 227 | 0.4892 | 0.8917 | 0.8719 | -0.1957 | -0.0973 |
| 1989.NL.Eng.CamioJoseCela | 5826 | 1541 | 0.2645 | 0.8058 | 0.9200 | -0.0364 | 0.2301 |
| 1990.BS.Eng.OctavioPaz | 636 | 300 | 0.4717 | 0.8686 | 0.8820 | -0.1903 | -0.0978 |
| 1990.NL.Eng.OctavioPaz | 5704 | 1549 | 0.2716 | 0.7870 | 0.9619 | -0.1224 | 0.1249 |
| 1991.BS.Eng.NadineGordimer | 562 | 280 | 0.4982 | 0.8924 | 0.8368 | -0.1779 | -0.1242 |
| 1991.GeorgeBush | 1771 | 582 | 0.3286 | 0.8438 | 0.8640 | -0.0115 | 0.0453 |
| 1991.NL.Eng.NadineGordimer | 4384 | 1254 | 0.2860 | 0.8021 | 0.9389 | -0.0878 | 0.1359 |
| 1992.BS.Eng.DerekWalcott | 104 | 68 | 0.6538 | 0.9300 | 0.8842 | -0.1905 | -0.2184 |
| 1992.NL.Eng.DerekWalcott | 7403 | 1965 | 0.2654 | 0.7743 | 0.9897 | -0.1805 | 0.1731 |
| 1993.BS.Eng.ToniMorrison | 368 | 201 | 0.5462 | 0.9122 | 0.7638 | -0.1078 | -0.1185 |
| 1993.MayaAngelou | 794 | 311 | 0.3917 | 0.8349 | 0.9974 | -0.2499 | -0.0170 |
| 1993.NL.Eng.ToniMorrison | 3486 | 1023 | 0.2935 | 0.8116 | 0.9188 | 0.0002 | 0.0533 |
| 1993.SarahBrady | 949 | 316 | 0.3330 | 0.8697 | 0.7814 | 0.1311 | 0.1014 |
| 1993.UrvashiVaid | 1315 | 414 | 0.3148 | 0.8399 | 0.8832 | 0.0114 | 0.0994 |
| 1994.MotherTeresa | 3953 | 638 | 0.1614 | 0.8150 | 0.9124 | 0.5166 | 0.7301 |
| 1994.NelsonMandela | 1010 | 388 | 0.3842 | 0.8479 | 0.8798 | -0.0936 | -0.0319 |
| 1995.BS.Eng.SeamusHeaney | 287 | 161 | 0.5610 | 0.9150 | 0.7679 | -0.1065 | -0.1995 |
| 1995.ErikaJong | 2356 | 601 | 0.2551 | 0.8298 | 0.8624 | 0.1873 | 0.2864 |
| 1995.HillaryClinton | 2483 | 715 | 0.2880 | 0.8225 | 0.8776 | 0.0696 | 0.1309 |
| 1995.NL.Eng.SeamusHeaney | 7050 | 1897 | 0.2691 | 0.7871 | 0.9577 | -0.1276 | 0.1895 |
| 1997.BillClinton | 1302 | 417 | 0.3203 | 0.8451 | 0.8357 | 0.1107 | 0.0794 |
| 1997.EarlOfSpencer | 1327 | 508 | 0.3828 | 0.8574 | 0.8216 | -0.0041 | -0.0359 |
| 1997.NancyBirdsall | 2312 | 644 | 0.2785 | 0.8328 | 0.8783 | 0.0783 | 0.2596 |
| 1997.PrincessDiana | 1753 | 602 | 0.3434 | 0.8498 | 0.8286 | 0.0174 | 0.1282 |
| 1997.QueenElizabethII | 449 | 207 | 0.4610 | 0.8998 | 0.7511 | -0.0231 | -0.0017 |
| 1999.AnitaRoddick | 2012 | 634 | 0.3151 | 0.8420 | 0.8630 | 0.0101 | 0.1252 |
| 2000.CondoleezzaRice | 1510 | 517 | 0.3424 | 0.8540 | 0.8794 | -0.0503 | 0.0159 |
| 2000.CourtneyLove | 8166 | 1627 | 0.1992 | 0.8000 | 0.9275 | 0.1890 | 0.3911 |
| 2000.PopeJohnPaulII | 823 | 327 | 0.3973 | 0.8747 | 0.8438 | -0.0937 | 0.0713 |
| 2001.09.11.GeorgeWBush | 670 | 296 | 0.4418 | 0.8807 | 0.7944 | -0.0727 | -0.0107 |
| 2001.09.13.GeorgeWBush | 550 | 251 | 0.4564 | 0.8744 | 0.8626 | -0.1634 | 0.0185 |
| 2001.BS.Eng.VSNaipaul | 347 | 172 | 0.4957 | 0.8982 | 0.8301 | -0.1708 | -0.0516 |
| 2001.HalleBerry | 636 | 219 | 0.3443 | 0.8496 | 0.8505 | 0.0575 | -0.0875 |
| 2001.NL.Eng.VSNaipaul | 6303 | 1220 | 0.1936 | 0.7873 | 0.9287 | 0.2345 | 0.4531 |
| 2002.OprahWinfrey | 603 | 226 | 0.3748 | 0.8653 | 0.8559 | -0.0308 | 0.0252 |
| 2003.BethChapman | 877 | 334 | 0.3808 | 0.8764 | 0.8497 | -0.0435 | 0.1807 |
| 2003.BS.Eng.JMCoetzee | 329 | 151 | 0.4590 | 0.9138 | 0.7498 | -0.0398 | 0.0728 |
| 2003.NL.Eng.JMCoetzee | 4593 | 1105 | 0.2406 | 0.7934 | 0.9693 | -0.0467 | 0.1642 |
| 2005.NL.Eng.HaroldPinter | 5787 | 1478 | 0.2554 | 0.8023 | 0.9161 | -0.0002 | 0.2048 |
| 2005.SteveJobs | 2584 | 706 | 0.2732 | 0.8321 | 0.8750 | 0.0857 | 0.2561 |
| 2007.NL.Eng.DorisLessing | 5881 | 1244 | 0.2115 | 0.7924 | 0.9689 | 0.0405 | 0.3426 |
| 2010.BS.Eng.MarioVargasLlosa | 452 | 204 | 0.4513 | 0.8911 | 0.7897 | -0.0722 | -0.0271 |
| 2010.NL.Eng.MarioVargasLlosa | 7073 | 2034 | 0.2876 | 0.7740 | 1.0218 | -0.2736 | 0.0711 |
| ErnestHemingway.TheOldManAndThe | 14915 | 1804 | 0.1210 | 0.7438 | 1.0362 | 0.3077 | 0.6836 |
| ErnestHemingway.TheOldManAndThe | 16130 | 1775 | 0.1100 | 0.7385 | 1.0420 | 0.3865 | 0.8258 |
| ErnestHemingway.TheSunAlsoRises.E | 22951 | 2099 | 0.0915 | 0.7049 | 1.0839 | 0.4104 | 0.7308 |
| IsaacAsimov.IRobot.Cap2 | 8772 | 1636 | 0.1865 | 0.7678 | 0.9552 | 0.1894 | 0.3930 |
| IsaacAsimov.IRobot.Cap6 | 12812 | 1977 | 0.1543 | 0.7597 | 0.9808 | 0.2709 | 0.5145 |

Spanish texts (1/3):
http://www.gfebres.com/F0IndexFrame/F132Body/F132BodyPublications/NatArtifLangs/Whole/Spanish.Properties.htm

## Spanish texts properties

***L*** Text Length
***D*** Diversity
***d*** Specific diversity [0-1]
***h*** Entropy [0-1]
***g*** Zipf's exponent
$J_{1,D}$ Zipf's diviation
$J_{\vartheta,D}$ Tail Zipf's diviation

| Text Name | *L* | *D* | *d* | *h* | *g* | $J_{1,D}$ | $J_{\vartheta,D}$ |
|---|---|---|---|---|---|---|---|
| 1755.PatrickHenry | 313 | 151 | 0.482 | 0.910 | 0.817 | -0.131 | -0.057 |
| 1805.Simón Bolívar | 462 | 230 | 0.498 | 0.878 | 0.942 | -0.190 | -0.196 |
| 1813.Simón Bolívar | 739 | 332 | 0.449 | 0.864 | 0.835 | -0.086 | -0.122 |
| 1819.Simón Bolívar | 11502 | 2629 | 0.229 | 0.751 | 0.962 | -0.030 | 0.226 |
| 1830.Simón Bolívar | 201 | 121 | 0.602 | 0.930 | 0.745 | -0.112 | -0.120 |
| 1863.AbrahamLincoln | 305 | 152 | 0.498 | 0.923 | 0.760 | -0.115 | 0.010 |
| 1868.CarlosMCespedes | 1457 | 591 | 0.406 | 0.836 | 0.945 | -0.214 | -0.082 |
| 1873.SusanBAnthony | 594 | 237 | 0.399 | 0.870 | 0.829 | -0.005 | -0.042 |
| 1899.Vladimir Lenin | 1920 | 644 | 0.335 | 0.817 | 0.948 | -0.140 | -0.002 |
| 1912.Emiliano Zapata | 2590 | 935 | 0.361 | 0.811 | 0.974 | -0.251 | -0.051 |
| 1917.Emiliano Zapata | 1619 | 653 | 0.403 | 0.826 | 0.985 | -0.260 | -0.094 |
| 1918.Emiliano Zapata | 1438 | 593 | 0.412 | 0.830 | 0.987 | -0.243 | -0.131 |
| 1918.WoodrowWilson | 303 | 171 | 0.564 | 0.909 | 0.813 | -0.158 | -0.160 |
| 1919.Georges Clemenceau | 209 | 126 | 0.603 | 0.928 | 0.792 | -0.179 | -0.174 |
| 1919.Lloyd George | 135 | 92 | 0.681 | 0.950 | 0.746 | -0.191 | -0.209 |
| 1921.MarieCurie.Esp | 563 | 239 | 0.425 | 0.895 | 0.785 | -0.010 | 0.022 |
| 1931.Manuel Azaña | 297 | 152 | 0.512 | 0.906 | 0.832 | -0.153 | -0.118 |
| 1933.JAntonioPrimoDeRivera | 3190 | 972 | 0.305 | 0.803 | 0.963 | -0.158 | 0.099 |
| 1934.Adolf Hitler | 347 | 163 | 0.470 | 0.893 | 0.870 | -0.146 | -0.100 |
| 1936.Dolores Ibarruri | 537 | 193 | 0.359 | 0.864 | 0.807 | 0.138 | -0.061 |
| 1936.José Buenaventura Durruti | 690 | 305 | 0.442 | 0.877 | 0.796 | -0.049 | -0.065 |
| 1938.Dolores Ibarruri | 774 | 318 | 0.411 | 0.846 | 0.962 | -0.218 | -0.042 |
| 1938.Leon Trotsky | 1023 | 416 | 0.407 | 0.860 | 0.835 | -0.063 | -0.044 |
| 1938.Neville Chamberlain | 638 | 302 | 0.473 | 0.883 | 0.827 | -0.131 | -0.077 |
| 1940.B.Winston Churchill | 68 | 36 | 0.529 | 0.892 | 1.048 | -0.136 | -0.191 |
| 1940.Benito Mussolini | 736 | 338 | 0.459 | 0.868 | 0.924 | -0.221 | -0.115 |
| 1940.Charles de Gaulle | 122 | 69 | 0.566 | 0.928 | 0.756 | -0.098 | -0.153 |
| 1940.Winston Churchill | 395 | 195 | 0.494 | 0.900 | 0.819 | -0.123 | -0.046 |
| 1941.Franklin Roosevelt | 280 | 158 | 0.564 | 0.922 | 0.803 | -0.151 | -0.131 |
| 1941.Joseph Stalin | 880 | 341 | 0.388 | 0.878 | 0.832 | -0.021 | 0.036 |
| 1942.08.Mahatma Gandhi | 2588 | 864 | 0.334 | 0.825 | 0.864 | -0.025 | 0.049 |
| 1943.Heinrich Himmler | 350 | 184 | 0.526 | 0.919 | 0.806 | -0.197 | -0.048 |
| 1943.Joseph Goebbels | 1173 | 414 | 0.353 | 0.864 | 0.798 | 0.101 | 0.031 |
| 1945.Harry Truman | 768 | 315 | 0.410 | 0.869 | 0.890 | -0.110 | -0.125 |
| 1945.Hirohito | 766 | 352 | 0.460 | 0.868 | 0.871 | -0.163 | -0.144 |
| 1945.Juan Domingo Perón | 1059 | 414 | 0.391 | 0.865 | 0.851 | -0.069 | 0.053 |
| 1946.Jorge Eliécer Gaitán | 3544 | 986 | 0.278 | 0.811 | 0.893 | 0.039 | 0.150 |
| 1947.George Marshall | 754 | 328 | 0.435 | 0.867 | 0.871 | -0.128 | -0.051 |
| 1948.David Ben Gurion | 1178 | 417 | 0.354 | 0.826 | 0.977 | -0.149 | -0.064 |
| 1950.Robert Schuman | 998 | 385 | 0.386 | 0.840 | 0.956 | -0.211 | -0.076 |
| 1950.William Faulkner | 533 | 233 | 0.437 | 0.895 | 0.761 | -0.006 | -0.002 |
| 1952.Eva Perón | 1124 | 344 | 0.306 | 0.839 | 0.899 | 0.030 | 0.071 |
| 1953.Dwight D Eisenhower | 1732 | 622 | 0.359 | 0.847 | 0.886 | -0.115 | 0.058 |
| 1956.Gamar Abdel Nasser | 839 | 337 | 0.402 | 0.868 | 0.821 | -0.050 | -0.059 |
| 1959.Fidel Castro | 2892 | 853 | 0.295 | 0.810 | 0.908 | -0.012 | 0.090 |
| 1959.Fulgencio Batista | 85 | 58 | 0.682 | 0.947 | 0.783 | -0.168 | -0.172 |
| 1959.Nikita Kruschev | 404 | 198 | 0.490 | 0.889 | 0.870 | -0.161 | -0.123 |
| 1961.J F Kennedy | 1613 | 602 | 0.373 | 0.838 | 0.894 | -0.136 | 0.014 |
| 1961.Nelson Mandela | 5350 | 1373 | 0.257 | 0.778 | 0.945 | -0.032 | 0.198 |
| 1962.J F Kennedy | 319 | 160 | 0.502 | 0.905 | 0.836 | -0.168 | -0.061 |
| 1963.J F Kennedy | 651 | 256 | 0.393 | 0.891 | 0.738 | 0.070 | 0.077 |

Spanish texts (cont. 2/3):

| | | | | | | | |
|---|---|---|---|---|---|---|---|
| 1963.Martin Luther King Jr | 1746 | 578 | 0.331 | 0.828 | 0.945 | -0.114 | 0.077 |
| 1964.Ernesto Che Guevara | 7172 | 1911 | 0.266 | 0.779 | 0.961 | -0.135 | 0.168 |
| 1964.Malcom X | 824 | 321 | 0.390 | 0.877 | 0.776 | 0.028 | 0.034 |
| 1964.Nelson Mandela | 5347 | 1372 | 0.257 | 0.778 | 0.944 | -0.032 | 0.198 |
| 1964.Ronald Reagan | 1062 | 450 | 0.424 | 0.875 | 0.789 | -0.049 | -0.026 |
| 1967.BS.Esp.MiguelAngelAsturias | 804 | 339 | 0.422 | 0.845 | 0.959 | -0.203 | -0.127 |
| 1967.Ernesto Che Guevara | 5868 | 1696 | 0.289 | 0.788 | 0.937 | -0.120 | 0.135 |
| 1967.Fidel Castro | 5519 | 1232 | 0.223 | 0.788 | 0.953 | 0.019 | 0.304 |
| 1967.Martin Luther King | 7418 | 1924 | 0.259 | 0.786 | 0.944 | -0.067 | 0.197 |
| 1967.NL.Esp.MiguelAngelAsturias | 4901 | 1533 | 0.313 | 0.787 | 0.967 | -0.184 | 0.038 |
| 1969.Richard Nixon | 4501 | 1200 | 0.267 | 0.800 | 0.925 | -0.026 | 0.210 |
| 1970.Salvador Allende | 1865 | 718 | 0.385 | 0.834 | 0.898 | -0.147 | -0.044 |
| 1971.BS.Esp.PabloNeruda | 468 | 209 | 0.447 | 0.859 | 0.946 | -0.193 | -0.120 |
| 1971.Pablo Neruda | 3683 | 1290 | 0.350 | 0.806 | 0.948 | -0.223 | -0.019 |
| 1972.Salvador Allende | 10046 | 2540 | 0.253 | 0.766 | 0.971 | -0.141 | 0.228 |
| 1973.Augusto Pinochet | 4191 | 1318 | 0.314 | 0.797 | 0.935 | -0.121 | 0.040 |
| 1973.Bando Nro 5 | 801 | 366 | 0.457 | 0.860 | 0.925 | -0.209 | -0.143 |
| 1973.Salvador Allende | 700 | 314 | 0.449 | 0.868 | 0.893 | -0.174 | -0.093 |
| 1974.Richard Nixon | 741 | 302 | 0.408 | 0.879 | 0.775 | 0.009 | 0.002 |
| 1976.Jorge Videla | 604 | 264 | 0.437 | 0.875 | 0.916 | -0.183 | -0.034 |
| 1977.BS.Esp.VicenteAleixandre | 241 | 137 | 0.568 | 0.917 | 0.850 | -0.209 | -0.134 |
| 1977.NL.Esp.VicenteAleixandre | 2379 | 859 | 0.361 | 0.818 | 0.988 | -0.253 | -0.010 |
| 1978.Juan Carlos I | 973 | 411 | 0.422 | 0.848 | 0.925 | -0.188 | -0.092 |
| 1979.Adolfo Suárez | 13201 | 2799 | 0.212 | 0.751 | 0.990 | -0.102 | 0.407 |
| 1979.Ayatolá Jomeini | 254 | 126 | 0.496 | 0.918 | 0.762 | -0.049 | -0.106 |
| 1979.Fidel Castro | 12832 | 2668 | 0.208 | 0.743 | 0.989 | -0.049 | 0.345 |
| 1981.Adolfo Suárez | 1348 | 420 | 0.312 | 0.842 | 0.818 | 0.088 | 0.099 |
| 1981.Roberto Eduardo Viola | 3823 | 1288 | 0.337 | 0.799 | 0.929 | -0.174 | 0.043 |
| 1982.BS.Esp.GabrielGarciaMarquez | 522 | 251 | 0.481 | 0.876 | 0.892 | -0.157 | -0.118 |
| 1982.Felipe González | 6592 | 1818 | 0.276 | 0.782 | 0.940 | -0.099 | 0.192 |
| 1982.Gabriel García Márquez | 2095 | 856 | 0.409 | 0.831 | 0.949 | -0.242 | -0.073 |
| 1982.Leopoldo Galtieri | 119 | 76 | 0.639 | 0.934 | 0.896 | -0.243 | -0.130 |
| 1982.Margaret Thatcher | 586 | 242 | 0.413 | 0.895 | 0.776 | -0.003 | 0.034 |
| 1983.Raúl Alfonsín | 3309 | 976 | 0.295 | 0.805 | 0.896 | 0.010 | 0.094 |
| 1984.Ronald Reagan | 790 | 339 | 0.429 | 0.864 | 0.825 | -0.073 | -0.059 |
| 1986.Ronald Reagan | 729 | 323 | 0.443 | 0.879 | 0.862 | -0.153 | -0.091 |
| 1987.Camilo José Cela | 1591 | 621 | 0.390 | 0.830 | 0.944 | -0.205 | -0.092 |
| 1987.Ronald Reagan | 3150 | 1016 | 0.323 | 0.816 | 0.924 | -0.144 | 0.083 |
| 1988.Gorbachov | 1017 | 416 | 0.409 | 0.859 | 0.847 | -0.093 | -0.085 |
| 1989.Carlos Saúl Menem | 1199 | 404 | 0.337 | 0.845 | 0.864 | 0.008 | 0.067 |
| 1989.NL.Esp.CamiloJoseCela | 6291 | 1803 | 0.287 | 0.777 | 0.965 | -0.148 | 0.085 |
| 1990.BS.Esp.OctavioPaz | 613 | 284 | 0.463 | 0.878 | 0.850 | -0.131 | -0.109 |
| 1990.George H. W. Bush | 654 | 269 | 0.411 | 0.881 | 0.854 | -0.114 | 0.049 |
| 1990.NL.Esp.OctavioPaz | 4804 | 1452 | 0.302 | 0.788 | 0.933 | -0.076 | 0.050 |
| 1991.Boris Yeltsin | 466 | 219 | 0.470 | 0.889 | 0.865 | -0.160 | -0.070 |
| 1991.Gorbachov | 197 | 126 | 0.640 | 0.936 | 0.715 | -0.133 | -0.161 |
| 1992.Rafael Caldera | 2504 | 832 | 0.332 | 0.810 | 0.932 | -0.150 | 0.048 |
| 1992.Severn Suzuki | 1001 | 403 | 0.403 | 0.869 | 0.876 | -0.157 | 0.040 |
| 1993.Bill Clinton | 2010 | 703 | 0.350 | 0.827 | 0.914 | -0.098 | -0.002 |
| 1996.Jose María Aznar | 5069 | 1383 | 0.273 | 0.782 | 0.951 | -0.104 | 0.197 |
| 1998.José Saramago | 6235 | 1775 | 0.285 | 0.781 | 0.978 | -0.182 | 0.106 |
| 1999.Elie Wiesel | 806 | 328 | 0.407 | 0.854 | 0.971 | -0.218 | -0.068 |
| 1999.Hugo Chavez | 12766 | 2441 | 0.191 | 0.760 | 1.002 | -0.051 | 0.442 |
| 2000.Vicente Fox | 7417 | 1998 | 0.269 | 0.778 | 0.929 | -0.066 | 0.173 |
| 2001.Fernando de la Rúa | 1129 | 436 | 0.386 | 0.853 | 0.825 | -0.042 | 0.004 |
| 2001.George W. Bush | 340 | 173 | 0.509 | 0.905 | 0.808 | -0.155 | -0.049 |
| 2001.Osama Bin Laden | 455 | 215 | 0.473 | 0.891 | 0.801 | -0.105 | -0.062 |
| 2002.A.George W. Bush | 590 | 271 | 0.459 | 0.887 | 0.820 | -0.095 | -0.032 |
| 2002.Barack Hussein Obama | 983 | 379 | 0.386 | 0.840 | 0.900 | -0.068 | -0.073 |
| 2003.B.George W. Bush | 564 | 237 | 0.420 | 0.886 | 0.823 | -0.056 | 0.013 |
| 2003.George W. Bush | 741 | 352 | 0.475 | 0.879 | 0.854 | -0.164 | -0.140 |
| 2003.José Saramago | 1110 | 441 | 0.397 | 0.849 | 0.870 | -0.083 | -0.062 |

Spanish texts (cont. 3/3):

| | | | | | | | |
|---|---|---|---|---|---|---|---|
| 2004.Pilar Manjón | 209 | 118 | 0.565 | 0.917 | 0.867 | -0.209 | -0.065 |
| 2005.Daniel Ortega | 7593 | 1516 | 0.200 | 0.779 | 0.943 | 0.125 | 0.347 |
| 2005.Gerhard Schroeder | 1547 | 559 | 0.361 | 0.843 | 0.875 | -0.076 | -0.011 |
| 2005.Steve Jobs | 2524 | 832 | 0.330 | 0.831 | 0.880 | -0.061 | 0.077 |
| 2006.Alvaro Uribe | 4555 | 1552 | 0.341 | 0.776 | 0.969 | -0.244 | -0.030 |
| 2006.Dianne Feinstein | 1503 | 525 | 0.349 | 0.841 | 0.888 | -0.088 | 0.018 |
| 2006.Evo Morales | 3391 | 890 | 0.262 | 0.812 | 0.981 | -0.154 | 0.287 |
| 2006.Gastón Acurio | 4348 | 1276 | 0.293 | 0.803 | 0.953 | -0.148 | 0.122 |
| 2006.Hugo Chavez | 3353 | 948 | 0.283 | 0.808 | 0.969 | -0.150 | 0.164 |
| 2007.Al Gore | 1319 | 580 | 0.440 | 0.859 | 0.903 | -0.228 | -0.097 |
| 2007.Cristina Kirchner | 5004 | 1228 | 0.245 | 0.795 | 0.918 | 0.047 | 0.262 |
| 2007.Daniel Ortega | 3373 | 857 | 0.254 | 0.805 | 0.969 | -0.082 | 0.282 |
| 2008.Barack Hussein Obama | 309 | 159 | 0.515 | 0.897 | 0.843 | -0.120 | -0.078 |
| 2008.J. L. Rodriguez Zapatero | 449 | 204 | 0.454 | 0.886 | 0.803 | -0.040 | -0.120 |
| 2008.Julio Cobos | 280 | 138 | 0.493 | 0.907 | 0.768 | -0.049 | -0.062 |
| 2008.Randy Paush | 1817 | 624 | 0.343 | 0.847 | 0.875 | -0.080 | 0.062 |
| 2009.Barack Hussein Obama | 2834 | 978 | 0.345 | 0.817 | 0.894 | -0.089 | -0.002 |
| 2010.BS.Esp.MarioVargasLlosa | 424 | 204 | 0.481 | 0.888 | 0.882 | -0.217 | -0.091 |
| 2010.Hillary Clinton | 2426 | 832 | 0.343 | 0.831 | 0.874 | -0.107 | 0.088 |
| 2010.NL.Esp.MarioVargasLlosa | 7034 | 2215 | 0.315 | 0.763 | 1.035 | -0.318 | 0.007 |
| 2010.Raúl Castro | 260 | 145 | 0.558 | 0.912 | 0.877 | -0.229 | -0.141 |
| 2010.Sebastian Piñera Echenique | 432 | 173 | 0.400 | 0.890 | 0.819 | -0.037 | 0.025 |
| CamiloJoseCela.LaColmena.Cap1 | 17409 | 3089 | 0.177 | 0.736 | 1.021 | 0.003 | 0.332 |
| CamiloJoseCela.LaColmena.Cap2 | 15370 | 2943 | 0.191 | 0.741 | 1.000 | -0.006 | 0.339 |
| CamiloJoseCela.LaColmena.Cap6 | 3629 | 1117 | 0.308 | 0.798 | 0.990 | -0.223 | 0.056 |
| CamiloJoseCela.LaColmena.Notas4E | 1623 | 596 | 0.367 | 0.829 | 0.954 | -0.171 | -0.031 |
| ErnestHemingway.ElViejoYElMar.Par | 13979 | 2498 | 0.179 | 0.751 | 0.975 | 0.116 | 0.452 |
| ErnestHemingway.ElViejoYElMar.Par | 15446 | 2424 | 0.157 | 0.743 | 0.993 | 0.186 | 0.542 |
| ErnestHemingway.Fiesta.Libro1 | 17642 | 3064 | 0.174 | 0.733 | 1.016 | 0.018 | 0.422 |
| GabrielGMarquez.CronMuerteAnunc | 12454 | 2621 | 0.210 | 0.754 | 0.948 | 0.080 | 0.248 |
| GabrielGMarquez.CronMuerteAnunc | 12680 | 2760 | 0.218 | 0.754 | 0.944 | 0.058 | 0.246 |
| GabrielGMarquez.CronMuerteAnunc | 6751 | 1586 | 0.235 | 0.774 | 0.933 | 0.088 | 0.193 |
| GabrielGMarquez.DicursoCartagena | 1443 | 579 | 0.401 | 0.844 | 0.910 | -0.175 | -0.081 |
| GabrielGMarquez.MejorOficioDelMu | 2949 | 1059 | 0.359 | 0.808 | 0.948 | -0.186 | -0.051 |
| IsaacAsimov.YoRobot.Cap2 | 8080 | 1856 | 0.230 | 0.767 | 0.967 | -0.020 | 0.220 |
| IsaacAsimov.YoRobot.Cap6 | 12235 | 2391 | 0.195 | 0.754 | 0.968 | 0.075 | 0.380 |
| JorgeLuisBorges.ElCongreso | 6656 | 1926 | 0.289 | 0.774 | 0.963 | -0.140 | 0.014 |
| JorgeLuisBorges.ElMuerto | 2109 | 753 | 0.357 | 0.814 | 0.950 | -0.174 | -0.067 |
| JorgeLuisBorges.ElSur | 2746 | 948 | 0.345 | 0.800 | 0.984 | -0.193 | -0.044 |
| JorgeLuisBorges.LasRuinasCirculares | 2238 | 824 | 0.368 | 0.826 | 0.920 | -0.138 | -0.046 |
| JoseSaramago.Valencia | 3711 | 1126 | 0.303 | 0.786 | 1.045 | -0.290 | 0.033 |
| MarioVargasLlosa.DiscursoBuenosAi | 1984 | 776 | 0.391 | 0.819 | 0.967 | -0.246 | -0.081 |
| MiguelAAsturias.SrPresidente.Parte | 4352 | 1269 | 0.292 | 0.786 | 0.975 | -0.143 | 0.057 |
| OctavioPaz.DiscursoZacatecas | 2238 | 711 | 0.318 | 0.810 | 0.949 | -0.101 | 0.013 |
| OctavioPaz.LaberintoSoledad.Part3 | 7054 | 1843 | 0.261 | 0.757 | 0.991 | -0.143 | 0.065 |

# Appendix B

## English: Diversity vs. Length

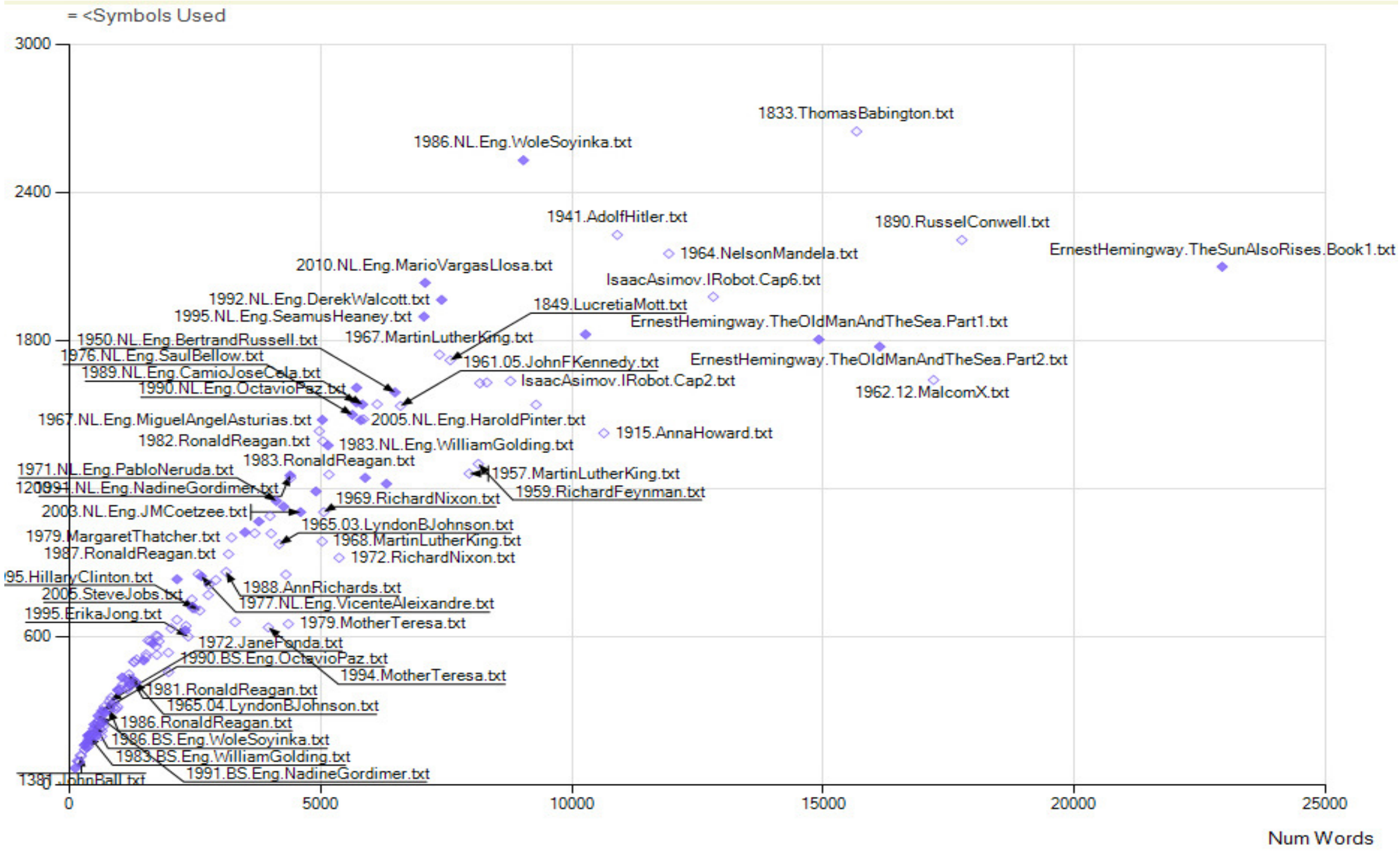

## Spanish: Diversity vs. Length

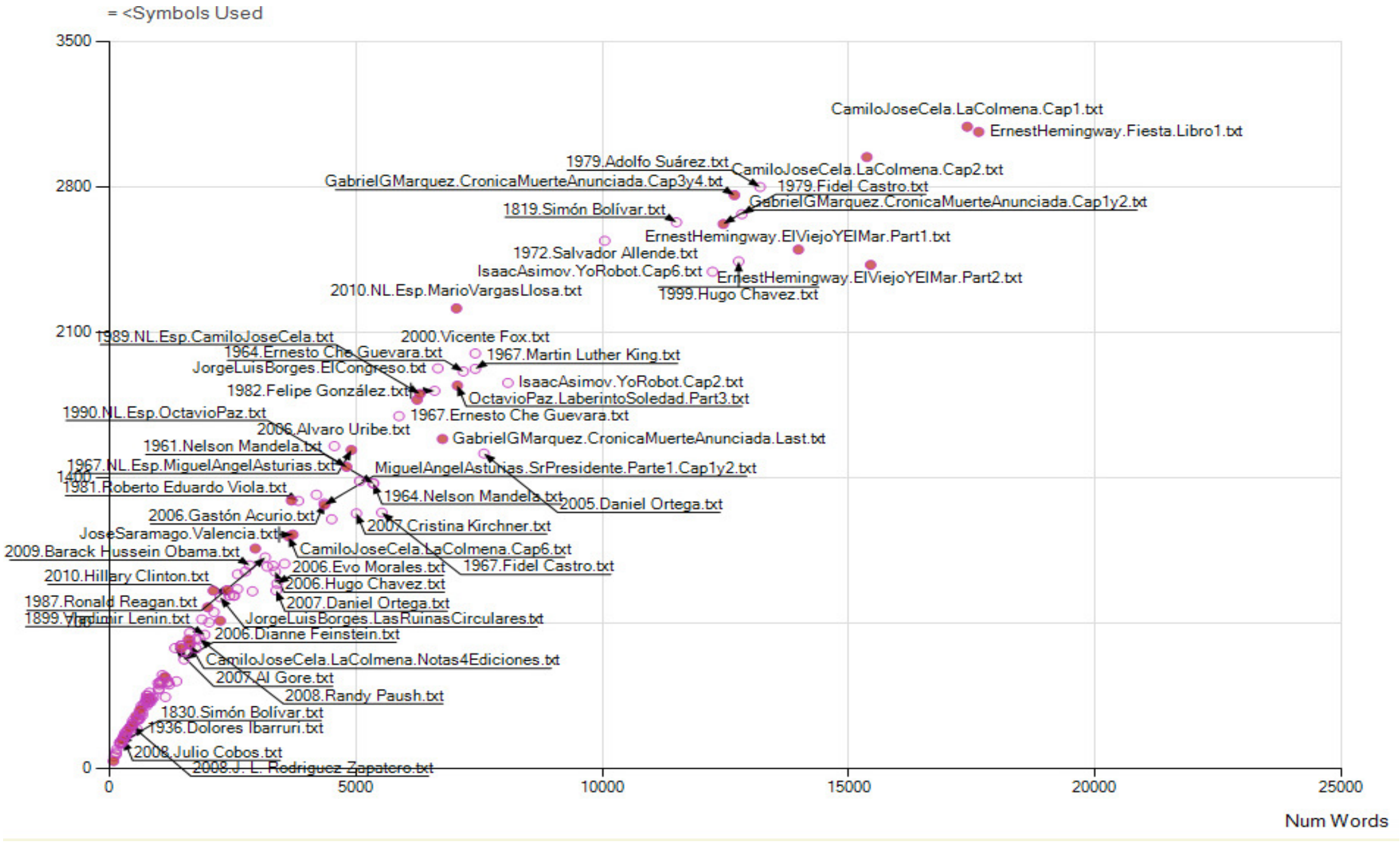